\pdfoutput=1

\documentclass[11pt]{article}

\usepackage[]{acl}

\usepackage{times}
\usepackage{latexsym}

\usepackage[T1]{fontenc}

\usepackage{graphicx}
\usepackage{tabu}
\usepackage{multirow}
\usepackage{enumitem}
\usepackage{amsmath}
\usepackage{booktabs}
\usepackage{hyperref}

\usepackage{diagbox}

\usepackage{subcaption}
\usepackage{amssymb}

\usepackage{microtype}

\usepackage{xcolor}
\colorlet{shadecolor}{blue!20}

\usepackage[utf8]{inputenc}

\title{Choose Your QA Model Wisely: A Systematic Study of Generative and Extractive Readers for Question Answering}

 \author{
    Man Luo \textsuperscript{\rm 1,2,}\thanks{~~Work done during internship at Salesforce Research.}
    \qquad Kazuma Hashimoto\textsuperscript{\rm 2}
    \qquad Semih Yavuz\textsuperscript{\rm 2}\\
    \textbf{
    \qquad Zhiwei Liu\textsuperscript{\rm 2} 
    \qquad Chitta Baral\textsuperscript{\rm 1}
    \qquad Yingbo Zhou\textsuperscript{\rm 2}} \\
    \textsuperscript{\rm 1} Arizona State University \textsuperscript{\rm 2} Salesforce Research  \\
    \textsuperscript{\rm 1} \texttt{\{mluo26, chitta\}@asu.edu}\\
    \textsuperscript{\rm 2}\texttt{\{k.hashimoto, syavuz, zhiweiliu, yingbo.zhou\}salesforce.com}
}
\begin{document}
\maketitle
\begin{abstract}
While both extractive and generative readers have been successfully applied to the Question Answering (QA) task, little attention has been paid toward the systematic comparison of them. 
Characterizing the strengths and weaknesses of the two 
readers is crucial not only for making a more informed reader selection in practice but also for developing a deeper understanding to foster further research on improving readers in a principled manner.
Motivated by this goal, we make the first attempt to systematically study the comparison of extractive and generative readers for question answering. 
To be aligned with the state-of-the-art, we explore nine transformer-based large pre-trained language models (PrLMs) as backbone architectures.
Furthermore, we organize our findings under two main categories: (1) keeping the architecture invariant, and (2) varying the underlying PrLMs.
Among several interesting findings, it is important to highlight that
(1) the generative readers perform better in long context QA, 
(2) the extractive readers perform better in short context while also showing better out-of-domain generalization, and
(3) 
the encoder of encoder-decoder PrLMs (e.g., T5) turns out to be a  strong extractive reader and outperforms the standard choice of encoder-only PrLMs (e.g., RoBERTa).
We also study the effect of multi-task learning on the two types of readers varying the underlying PrLMs and perform qualitative and quantitative diagnosis to provide further insights into future directions in modeling better readers.

 
\end{abstract}

\section{Introduction}

Question Answering (QA) is an important task to evaluate the reading comprehension capacity of an intelligent system and can be directly applied to real applications such as search engines~\citep{kwiatkowski-etal-2019-natural} and dialogue systems~\citep{reddy-etal-2019-coqa,choi-etal-2018-quac}.  This paper studies extractive QA which is a specific type of QA; i.e., answering the question using a span from the context~\citep{rajpurkar-etal-2016-squad,fisch-etal-2019-mrqa}. Extractive readers~\citep{Seo2017BidirectionalAF,devlin-etal-2019-bert} are widely used to tackle such a task, where the goal is to classify start and end positions of the answer in the context. Generative readers~\citep{Raffel2020ExploringTL,Lewis2020RetrievalAugmentedGF,izacard-grave-2021-leveraging} have also shown remarkable performance, where the goal is to generate answers by autoregressively predicting tokens.

Both the state-of-the-art extractive and generative readers are based on large pretrained language models (PrLMs) and show good performance on different datasets. However, a systematic comparison between them has been largely unexplored. Such a comparison reveals the strengths and weaknesses of each reader, which in turn can provide more principled guidance on which reader and PrLM should be applied in which cases, and also open up future research opportunities grounded on identified concrete challenges to improve reader models.
However fair comparisons between these have been difficult to perform mainly because 1) the PrLMs for extractive and generative are different, i.e., extractive readers are usually built on top of encoder-only PrLM while generative ones are based on encoder-decoder PrLMs, and 2) the size of generative and extractive readers are not the same, which can greatly affect the performance. 
We design two main set of controlled experiments to address such challenges in comparing extractive and generative readers in a principled manner.

In the first set of experiments, we compare extractive and generative readers using the same PrLMs. Specifically, T5~\citep{Raffel2020ExploringTL} generative reader is compared with T5 extractive reader and  similarly for BART~\citep{lewis2020bart}.
This allows a fair comparison of different answer prediction approaches without being affected by different architecture or prior knowledge of PrLMs.
Moreover, we challenge the conventional formulation of extractive readers, which are often built upon encoder-only PrLMs, by leveraging the encoder of encoder-decoder PrLMs as a variable alternative. More concretely, we use the encoders of T5 and BART models to explore their capacity as an extractive reader to better understand the effect of different pre-training strategies on the final QA performance.

While the aforementioned comparison strategy adopts the same PrLMs, it remains unclear how generative readers compare with the conventional extractive readers that are built upon encoder-only PrLMs. 
Thus, in the second experiment, we compare different architecture PrLMs, including T5, BART, ELECTRA~\cite{Clark2020ELECTRAPT} and RoBERTa~\cite{Liu2019RoBERTaAR}, to draw more generalizable and grounded conclusions. 
All models in this suite of experiments have similar sizes, thus reducing the impact of model size on performance.
With these two experiments, we present a systematic comparison of extractive and generative readers using nine readers on the MRQA task~\citep{fisch-etal-2019-mrqa}, a collection of multiple extractive QA datasets.
This evaluation results in five insightful findings:
\begin{enumerate}[nosep,noitemsep,leftmargin=*]
\item The first experiment reveals that the choice of PrLM affects the performance. Specifically, for T5, the generative reader is better than the extractive one, but for BART, extractive readers are better than the generative ones. 
\item The second experiment shows that on average, extractive readers performs better than the generative ones, with the extractive reader built on the encoder of T5 performing the best among the different types of PrLMs.
\item Extractive readers perform better in short context and have better generalization on out-of-domain datasets and rare answers, but the generative readers perform better in the long context.
\item  The encoder of encoder-decoder PrLMs are also good extractive readers. Extractive readers built on top of the encoder of BART or T5 are better than encoder-only PrLMs, like RoBERTa. 
\item While the inference length is usually chosen to be the same as in the training time, we find that longer inference length has a positive effect for all PrLMs. Using longer lengths for long contexts leads to greater gains than short contexts. 
\end{enumerate}
Our work presents an in-depth study of extractive and generative readers for QA task, an important NLP task toward building intelligent systems. 
Our findings shed light on key considerations behind reader selection and would be helpful for formulating future research on advancing reader models.
\section{Related Work}

\paragraph{Pretrained Language Models}
Here, we mainly discuss two types of pre-trained models based on transformers architecture~\citep{Vaswani2017AttentionIA}, autoencoder and encoder-decoder models, which are widely used for QA tasks. Autoencoder only relies on the encoder part in the original transformer, and in the pretraining time, the input is a corrupted sentence, for example, a sentence with mask tokens, such as BERT~\citep{devlin-etal-2019-bert} and RoBERTa ~\citep{Liu2019RoBERTaAR} and ELECTRA~\citep{Clark2020ELECTRAPT}. Both RoBERTa and ELECTRA has the same architecture as BERT but perform better than BERT on many tasks. RoBERTa mainly benefits from larger training corpus  consisting of news, books, stories,
and web text. ELECTRA adapts GAN-style training~\citep{Mirza2014ConditionalGA}
and aims to detect if a token is replaced or is from the original text. Large ELECTRA is trained on similar data as RoBERTa. BART~\citep{lewis-etal-2020-bart} and T5~\citep{Raffel2020ExploringTL} belong to encoder-decoder architecture. BART is pretrained on the same data as RoBERTa, while T5 is pre-trained on Colossal Clean Common Crawl Corpus as well as the multiple downstream tasks.

\paragraph{Question Answering Systems}

We focus on QA systems that are built upon PrLMs. Extractive QA readers assume that answers can be found in the context and aim to predict the corresponding start and end tokens from the context~\citep{fisch-etal-2019-mrqa,li-etal-2019-net,Clark2020ELECTRAPT,karpukhin-etal-2020-dense}. Differently, generative QA readers are not restricted to the input context, where they can freely generate answers token by token using the entire vocabulary in an autoregressive manner~\citep{Raffel2020ExploringTL}. Generative readers are more often used in open domain ~\cite{Lewis2020RetrievalAugmentedGF,izacard-grave-2021-leveraging,Xiong2021AnsweringCO} and unified settings~\citep{khashabi-etal-2020-unifiedqa,Tafjord2021GeneralPurposeQW}. ~\citet{Fajcik2021PruningTI} combines extractive and generative readers by adding a classification module to decide which reader predicts answers. \citet{cheng-etal-2021-unitedqa} proposes a unified system of extractive and generative readers, but different from \citep{Fajcik2021PruningTI}, the output is computed by both extractive and generative readers. 

\section{Model}\label{sec:model}
We mainly study the QA models based on PrLMs with extractive and generative approaches. 
\subsection{Extractive Reader} \label{sec:ext_model}

In extractive reader, an encoder firstly receives the concatenation of a question $\mathbf{q:}\{q_1, \dots, q_t\}$ and a context $\mathbf{c:}\{c_1, \dots, c_m\}$, where $q_i$ and $c_j$ are tokens in question and context, respectively. 
Then, it produces $\mathbf{h}: [ h_1 | \cdots | h_m ] \in \mathbb{R}^{d \times m}$, where $h_j$ corresponds to the $d$-dimensional contextual representation of context token $c_j$. We then stack two linear layers on top of the contextual representations to independently predict the probability of each context token being start and end positions of the correct answer. More formally, given a tuple $(\mathbf{q}, \mathbf{c}, \mathbf{a})$, where $\mathbf{a}$ is an answer, the training objective is to minimize the following loss function
\begin{align}
\mathcal{L}_{\text{Ext}} = -\log(\mathbf{P_{start,s}}) -\log(\mathbf{P_{end,e}})
\end{align}
where $\mathbf{P_{start}}, \mathbf{P_{end}} \in \mathbb{R}^{m}$ are defined by
\begin{align}
\mathbf{P_{start}} &= \text{softmax}(\mathbf{w_{start}}\mathbf{h})\\
\mathbf{P_{end}} &= \text{softmax}(\mathbf{w_{end}}\mathbf{h})
\end{align}
where $\mathbf{w_{start}}$ and $\mathbf{w_{end}}$ denote for the linear layers to predict start and end tokens, ${\mathbf{P_{start,s}}}$ and ${\mathbf{P_{end,e}}}$ denote the probability of the ground truth start and end tokens of answer $\mathbf{a}$, respectively.
In testing time, the answer span is decoded by $\text{argmax}_{i, j} \{{\mathbf{P_{start,i}}} \times {\mathbf{P_{end,j}}}\}$. 

In this work, we have two variants of extractive readers. One is encoder-only models to get the contextual representation of each token. We call such kind of reader as \textbf{E-Extractive reader}.  Apart from taking the conventional PrLMs such as RoBERTa and ELECTRA, we also apply the encoder part in T5 and BART to be E-Extractive reader. The other one is using the encoder-decoder models where the decoder is to obtained the contextual representation of each token in the context in an autoregressive way (see \S\ref{sec:gen_model}). We use both BART and T5 PrLMs and term this kind of reader as \textbf{ED-Extractive reader}.  



\subsection{Generative Reader} \label{sec:gen_model}
We consider a generative reader consisting of an encoder and a decoder where the decoder is used to generate answers in an autoregressive way. Specially, the encoder takes a question $\mathbf{q}$ and a context $\mathbf{c}$ as input and outputs contextual representation $\mathbf{h}$. 
Then, the decoder takes the previously generated answer tokens as input and performs attention over $\mathbf{h}$ and then generates the next token. Formally, given a tuple $(\mathbf{q}, \mathbf{c}, \mathbf{a})$, the training objective is to minimize the following loss function 
\begin{equation}
\mathcal{L}_{\text{Gen}} = \sum_{i=1}^{K}\log \mathbf{P}(a_i\mid \mathbf{h}, a_{:i})
\end{equation}
where $K$ is the number of tokens in answer $\mathbf{a}$, $a_i$ is the $i^{th}$ token in $\mathbf{a}$, and $a_0$ corresponds to a special beginning of sequence (\texttt{BOS}) token.
In the inference time, we use the greedy search method to autoregressively generate the answer.

\section{Experiments}
\subsection{Dataset} 
We conduct experiments on MRQA benchmark which  provides six in-domain (IID) datasets, and six out-of-domain (OOD) datasets for generalization evaluation. MRQA covers different domains (e.g. News and biomedical) and different types of questions, (e.g. single hop and multi-hop). Table \ref{tab:datasets} shows the statistic of each IID and OOD dataset. Some datasets have long context and others are short context. More details about MRQA are presented in Appendix \ref{apd:datasets}. 
\begin{table}[t]
\centering
 \resizebox{0.95\linewidth}{!}{
\begin{tabular}{l|c|c|c}
    \toprule
     {\bf Dataset}  &  {\bf Training size }  & {\bf Avg. tokens in Q} & {\bf  Avg. tokens in C}  \\
     \toprule
    \multicolumn{3}{l}{\bf In-domain datasets}\\
    \bottomrule
        SQuAD  &  86,588  & 11.53 & 144.15  \\
        NewsQA  &  74,160  & 7.60 & 581.61  \\
        TriviaQA  & 61,688  & 15.81 & 782.59  \\
        SearchQA  & 117,384  & 17.46 & 744.44  \\
        HotpotQA  & 72,928  & 18.89 & 237.67  \\
        NQ  & 104,071 & 9.18 & 158.80 \\
    \toprule
    \multicolumn{3}{l}{\bf Out-of-domain datasets}\\
    \bottomrule
        DROP  &  -  & 11.18 & 215.16  \\
        RACE  &  -  & 11.82 & 347.90  \\
        BioASQ  & -  & 11.53 & 252.83  \\
        TextbookQA  & -  & 11.07 & 663.36  \\
        RE  & -  & 9.26 & 30.02  \\
        DuoRC  & - & 8.63 & 732.92 \\
    \toprule
    \end{tabular}
    }
\caption{Statistics of In-domain (IID) and out-of-domain (OOD) datasets of MRQA benchmark.}
\label{tab:datasets}
\end{table}

\subsection {Learning Strategy}

{\bf Single Task Learning:} we use each IID datasets to train extractive and generative readers. {\bf Multi-Task Learning:} we consider training with all (six) IID datasets as multi-task learning for two reasons.  As \citep{su-etal-2019-generalizing} showed that different IID datasets share a low similarity, therefore, they may require different reasoning skills. In addition, Table \ref{tab:datasets} shows that different datasets have different question and context lengths, which may lead to different difficulties between datasets.


\subsection{Experimental Setup}
\begin{table*}[t]
\centering
 \resizebox{\linewidth}{!}{
\begin{tabular}{l|c|c|c|c|c|c|c|c|c}
    \toprule
     ~  &  {\bf T5 E-Ext } & {\bf T5 E-Ext } & {\bf T5 ED-Ext } & {\bf T5 ED-Gen } & {\bf Bart E-Ext} & {\bf Bart ED-Ext} & {\bf Bart ED-Gen} &{\bf ELECTRA}  & {\bf RoBERTa}  \\
     \toprule
        Size  &  base  & large & base  & base & large & large & large & large & large\\
        \midrule
        \# Params (M)  & 110  & 335 & 223  & 223  & 204 & 406 & 406 & 334 & 354\\
    \bottomrule
    \end{tabular}
    }
\caption{Size and parameters of readers. ED: encoder-decoder, Ext for extractive, Gen for generative approach.}
\label{tab:model_size}
\end{table*}
\begin{table}[t]
\large
\centering
\renewcommand{\arraystretch}{1.2}
{
\resizebox{\linewidth}{!}{
\begin{tabular}{c|lll|lll}
\toprule
\multirow{2}{*}{Model} &  \multicolumn{3}{c}{In-domain Avg.} & \multicolumn{3}{|c}{Out-of-domain Avg.}\\
\cmidrule(lr){2-4} \cmidrule(lr){5-7}
~& 512 & 1024 & Full &512 & 1024 & Full  \\ 
\toprule
\multicolumn{3}{l}{\bf Single Task Learning}\\
\bottomrule
T5 E-Ext (B)  & 74.42 & 75.80 & \textbf{77.93} &55.89 & 58.06 & \textbf{58.65}   \\ 
T5 E-Ext (L)  & 76.46 & 78.67 & \textbf{80.85} & 60.74 & 63.67 & \textbf{64.49}  \\ 

T5 ED-Ext (B)  & 74.75 & 77.06 & \textbf{79.11} & 57.11 & 59.19 & \textbf{59.99}  \\ 

T5 ED-Gen (B)  & 77.91 & 80.68 & \textbf{81.02} & 56.26 & 61.75 & \textbf{61.82}  \\ 
BART E-Ext (L) &  77.78 & \textbf{79.10} & - & 59.67 & \textbf{61.32} & -  \\ 
BART ED-Ext (L) & 77.10 & \textbf{77.34} & - & \textbf{59.29} & 59.21 & -  \\ 
BART ED-Gen (L) &  69.89 & \textbf{70.24} & - & 49.65 &\textbf{53.51} & -   \\ 
RoBERTa (L) & 77.59 & \textbf{77.89} & - & 60.32 & \textbf{60.47} & -  \\ 
ELECTRA (L) & 78.71 & - & - & 60.19 & - & -  \\ 

\toprule
\multicolumn{3}{l}{\bf Multi-Task Learning}\\
\bottomrule
T5 E-Ext (B)  &  75.74 & 76.65 & \textbf{78.99} &58.94 & 61.55 & \textbf{61.98}\\ 
T5 E-Ext (L)  & 77.10 & 79.30 & \textbf{81.55} &63.04 & 66.10 & \textbf{66.78}  \\ 

T5 ED-Ext (B)  & 75.92 & 77.38 & \textbf{79.93} & 59.23 & 61.86 & \textbf{62.64}  \\ 

T5 ED-Gen (B) & 78.06 & 80.89 & \textbf{81.16} &57.82 & 63.56 & \textbf{63.68}  \\ 
BART E-Ext (L) &  77.75 & \textbf{79.13} & - &63.27 & \textbf{64.06} & -  \\ 
BART ED-Ext (L)  & 77.26 & \textbf{77.55} & - & 62.14 & \textbf{62.68} & -  \\ 
BART ED-Gen (L) &  78.11 & \textbf{78.55} & - &57.41 & \textbf{60.54} & -   \\ 
RoBERTa  (L) & 77.86 & \textbf{78.02} & - &\textbf{63.70} & 63.58 & -  \\ 

ELECTRA  (L) & 78.52 & - & - &63.83 & - & -  \\ 

\bottomrule 
\end{tabular}
}
}
\caption{Result of each model using three inference length. \textbf{Bold} number means the highest value of each model with three inference length for IID and OOD datasets. L: large PrLMs, B: base PrLMs}
\label{tab:diff_infer_length}
\end{table}
We use Huggingface~\citep{wolf-etal-2020-transformers} and Pytorch~\citep{NEURIPS2019_9015} implementation for training each model. 
All models are trained using maximum input length of 512 and other details is provided in Appendix \ref{apd:setup}\footnote{While we fix the training hyperparameters for all the models for the sake of experimental efficiency, the performance of our setting is close to the original results.}. 
In Table \ref{tab:model_size}, we summarize the size of each evaluated model and the size of PrLMs are chosen based on a comparable way and the best computation power. 
For example, we choose T5 base model for generative reader since the large T5 is too larger (737M). 

\noindent\textbf{Input Format}:
Given a question {\bf Q} and a context {\bf C}, the input to extractive readers is \{{\bf Q} [SEP] {\bf C}\} and the input to generative readers is \{{\textit{question:} \bf Q} [SEP] \textit{context: }{\bf C}\}. 
We also considered other input formats, which are reported in Appendix \ref{apd:two_input_format}.
\noindent\textbf{Answer Length of Generative Reader}: We set the maximum generated answer length as 16 for generative reader. Using longer generation lengths (32 and 64) do not yield noticeable improvement as reported in Appendix \ref{apd:gen_ans_length}.

\section{Results and Analysis} \label{sec:result}
We first present the study of using different inference length for each model since it guides us to choose the best performance of each model. Then, we compare the generative and extractive readers using the same PrLMs and the different PrLMs. Last, we present a detail analysis to diagnose the difference among extractive and generative reader. 
F1 is used to measure performance. 
Note that since we test each model on 12 datasets, the observation and conclusion we draw are mostly based on the average across all datasets. 

\subsection{The Effect of Context Length} \label{sec:length_effect}

While all models are trained with $512$ maximum length, the inference length can be longer than this. We experiment with three lengths, $512$, $1024$, and the full length of input question and context. Due to the tokenization and pretraining maximum length of each PrLM, 
ELECTRA only allows $512$ maximum inference length, RoBERTa and BART allows $1024$, and T5 allows the full length of input. 

We present the average performance of each model on both IID and OOD in Table \ref{tab:diff_infer_length}\footnote{Note that in single-task learning, the performance on OOD are extracted from the best performance of each single-task model on every dataset and this applies to all other tables in this paper.}, from which three trends are observed. (1) When using $512$ inference length, ELECTRA is the best model in single-task learning on IID datasets and multi-task in both IID and OOD datasets. (2) Increasing the inference length actually improves all models' performance. (3) The length affects the T5 models more significantly than others, for example, in single-task learning, the largest improvement of length $1024$ for T5 model on IID and OOD datasets are  $2.77\%$ and $5.49\%$, while for other models, the largest improvement of length $1024$ compared to $512$ are $1.32\%$ and $1.65\%$. 
The performance of using 512 and 1024 are given in Appendix \ref{apd:infer_length}, and we present the performance of each dataset using the best input length in the following sections.

\subsection{Comparison within Same PrLMs} \label{sec:comp_same_prlms}




We compare different readers when using the same PrLMs. 
Two PrLMs, T5 and BART, are considered, where T5-base model is applied to each T5 reader, and BART-large model is applied to each BART reader.
We have three comparison as there are two types of extractive and one type of generative readers (\S\ref{sec:model}).
We present the average performance in each comparison and the detail performance on each datasets are given in Appendix \ref{apd:same_prlms}. 

\begin{figure}[h!]
\noindent\begin{subfigure}[b]{0.25\textwidth}
    \includegraphics[width=0.99\textwidth]{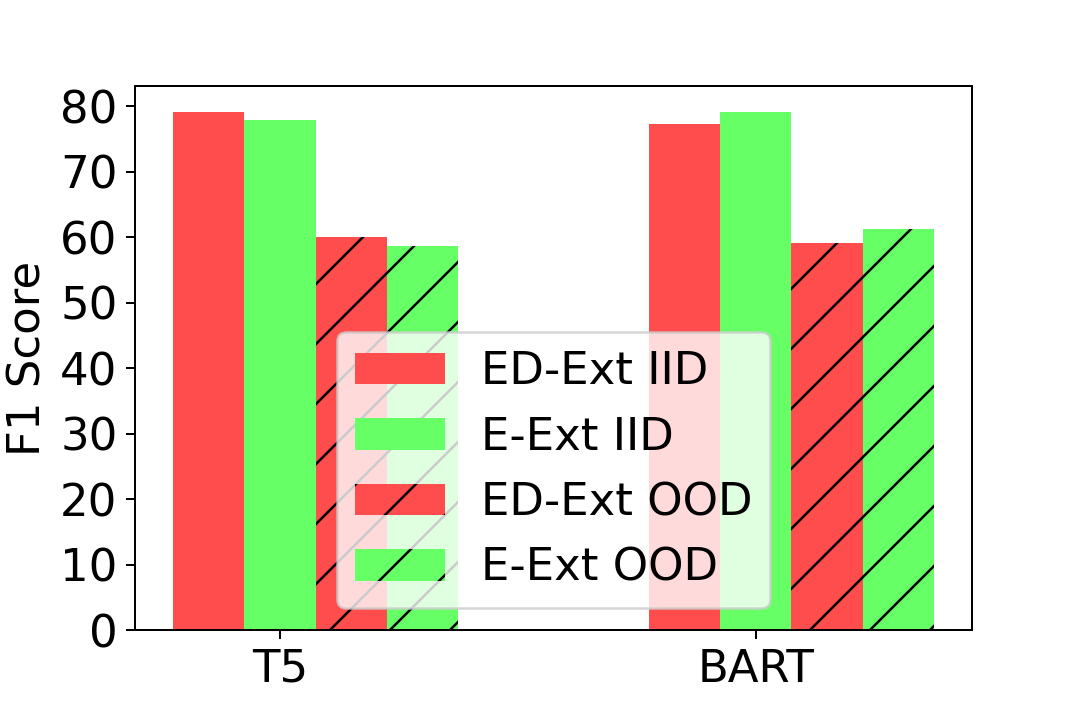}
\end{subfigure}%
\noindent\begin{subfigure}[b]{0.25\textwidth}
    \includegraphics[width=0.99\textwidth]{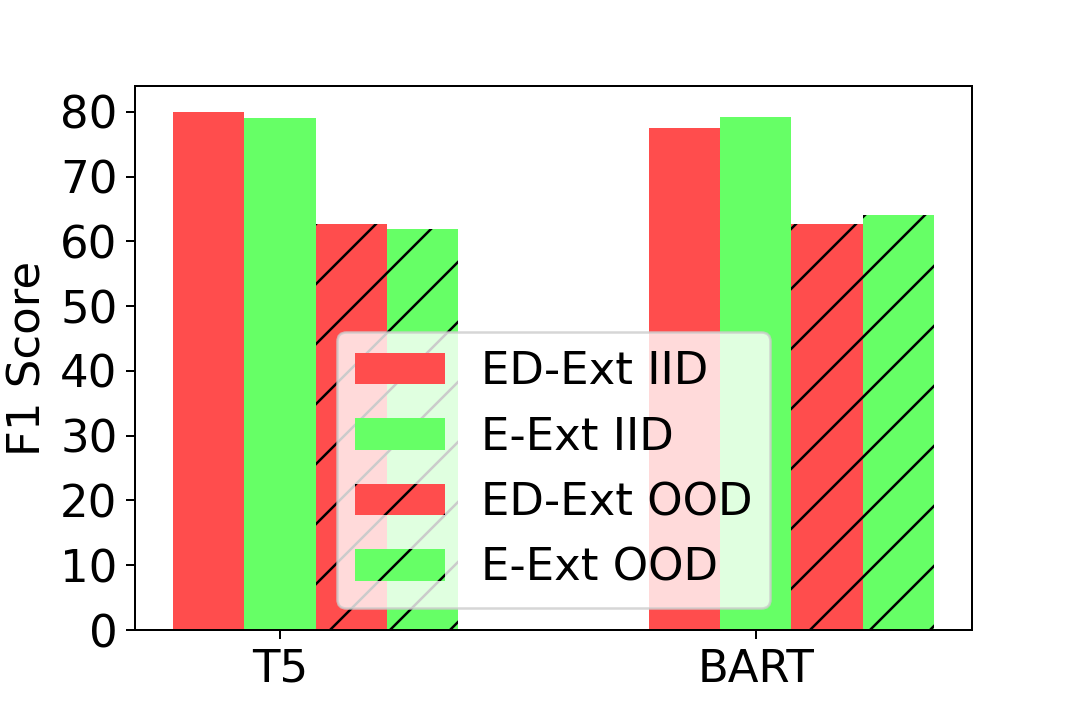}
\end{subfigure}%
\caption{
Left for single-task and right for multi-tasks settings.
For T5, ED-Ext performs better than E-Ext reader; for BART, E-Ext is better than ED-Ext reader even though the former has less parameters.}
\label{fig:compare_ed_e_ext}
\end{figure}

\paragraph{ED-Extractive and E-Extractive}
Since the E-Extractive reader is only use the encoder part of the PrML without the decoder, the size of E-Extractive reader is less than the ED-Extractive. But even under this disadvantage, surprisingly, we find that the encoder part actually perform well on QA tasks. 
In Figure \ref{fig:compare_ed_e_ext} , the \textcolor{black}{red} and \textcolor{black}{green} bars compare the  \textcolor{black}{ED-Extractive} and \textcolor{black}{E-Extractive} reader. 
For BART model, the E-Extractive reader outperforms ED-Extractive reader on average on IID and OOD datasets in single task learning as well as multi-task learning. This indicates that the decoder in BART is not crucial for the extractive reader. On the other hand, for T5, the ED-Extractive reader outperforms E-Extractive reader on average on both IID and OOD datasets. This suggests that the decoder in T5 still plays a role to yield better performance. But the performances are similar even that the E-Extractive reader has less parameters. 

\begin{figure}[h!]
\noindent\begin{subfigure}[b]{0.25\textwidth}
    \includegraphics[width=0.99\textwidth]{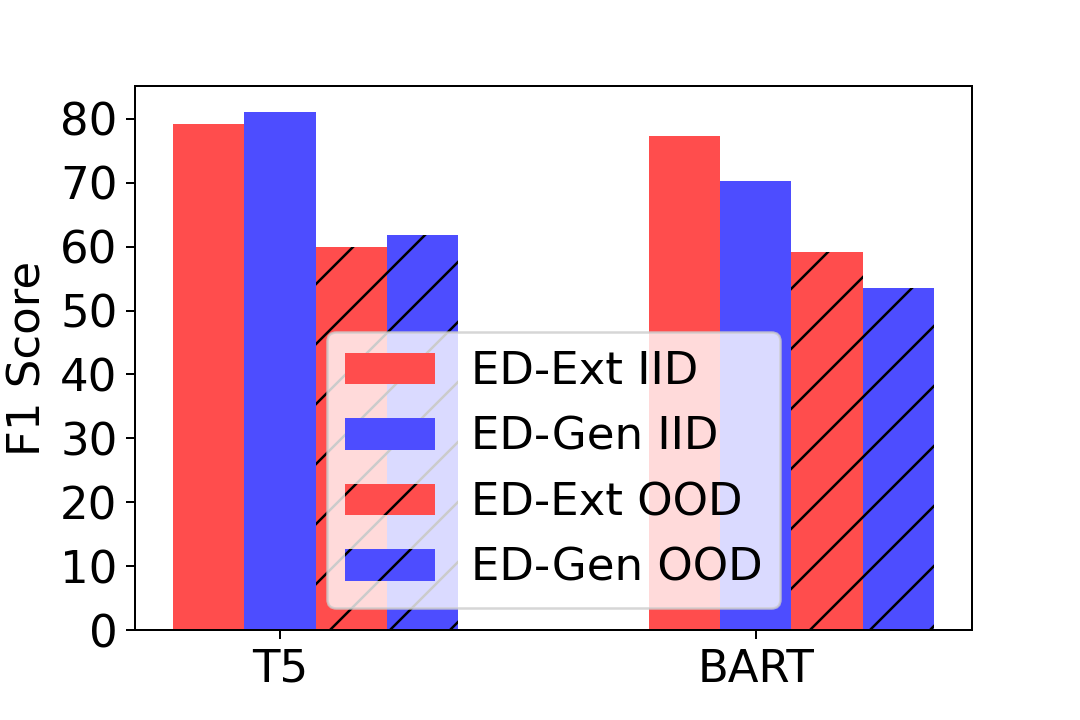}
\end{subfigure}%
\noindent\begin{subfigure}[b]{0.25\textwidth}
    \includegraphics[width=0.99\textwidth]{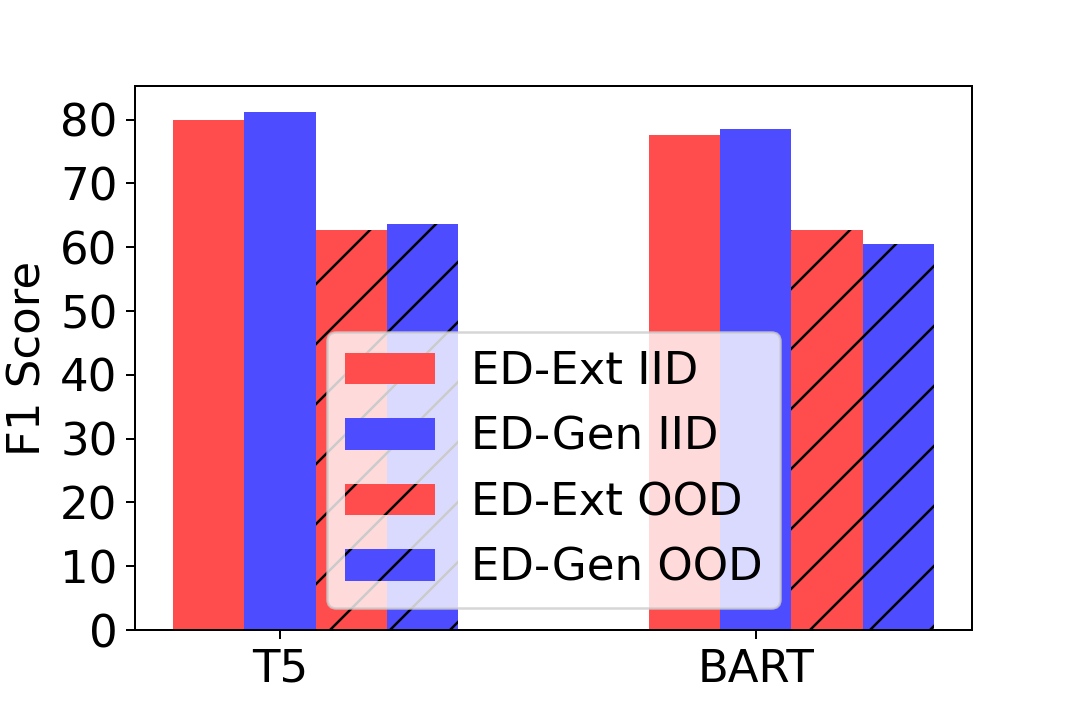}
\end{subfigure}%
\caption{
Left for single-task and right for multi-tasks settings.
For T5, ED-Gen performs better than ED-Ext; For BART, ED-Ext is better than ED-Gen in single task learning, but worse in multi-task learning.}
\label{fig:comp_ed_ext_gen}
\end{figure}

\paragraph{ED-Extractive and ED-Generative Reader} 
Here, the model size of extractive reader and generative reader are almost the same (see Table \ref{tab:model_size}) and also the pre-owned knowledge of two readers are the same since both readers use the encoder and decoder parts. In Figure \ref{fig:comp_ed_ext_gen}, the \textcolor{black}{red} and \textcolor{black}{blue} bars compare the \textcolor{black}{ED-Extractive} and \textcolor{black}{ED-Generative} reader. For T5, generative models performs better than the extractive one on four cases, IID and OOD datasets and single- and multi-tasks learning.
For BART PrLM, in single-task learning, the extractive model is much better than the generative model. This probably explains why in most of the previous work, when BART is applied to extractive QA tasks, it is used as extractive reader even though it belongs to encoder-decoder model family\footnote{The original BART paper takes BART as an extractive and also the implementation of using BART for QA in Huggingface library do the same.}. The story for multi-task learning is different, and we find that the BART generative reader benefits significantly from multi-task learning and even outperforms the BART ED-extractive reader on IID datasets. It indicates that the decoder in BART requires larger and more diversified datasets to learn the QA task.

\begin{figure}[h!]
\noindent\begin{subfigure}[b]{0.25\textwidth}
    \includegraphics[width=0.99\textwidth]{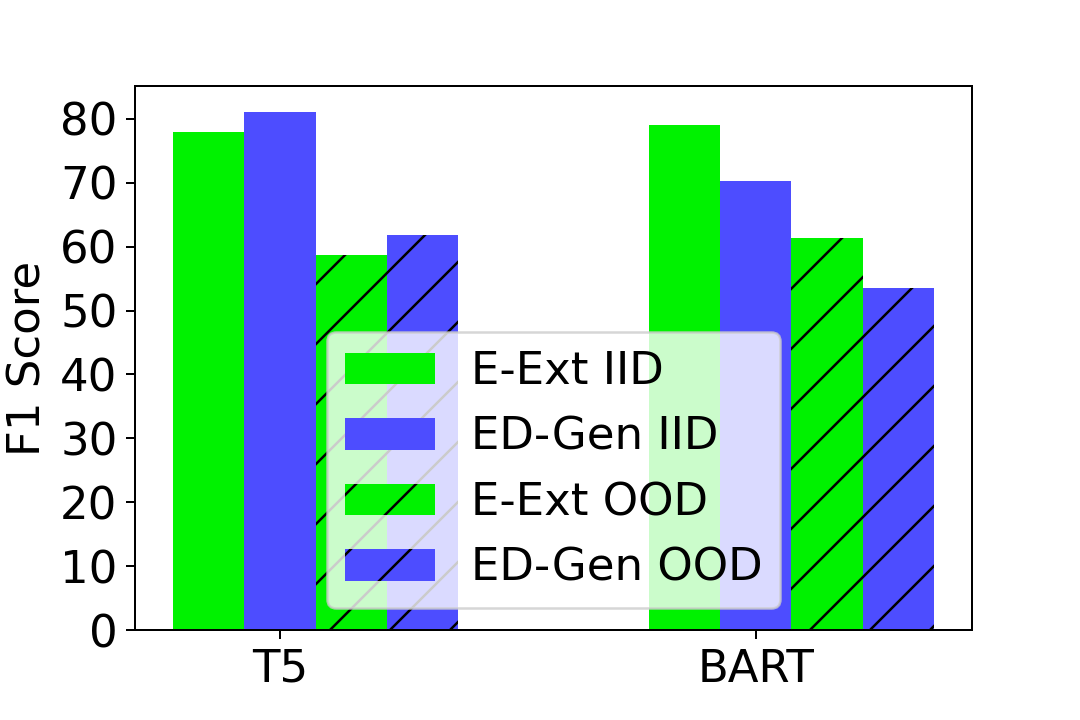}
\end{subfigure}%
\noindent\begin{subfigure}[b]{0.25\textwidth}
    \includegraphics[width=0.99\textwidth]{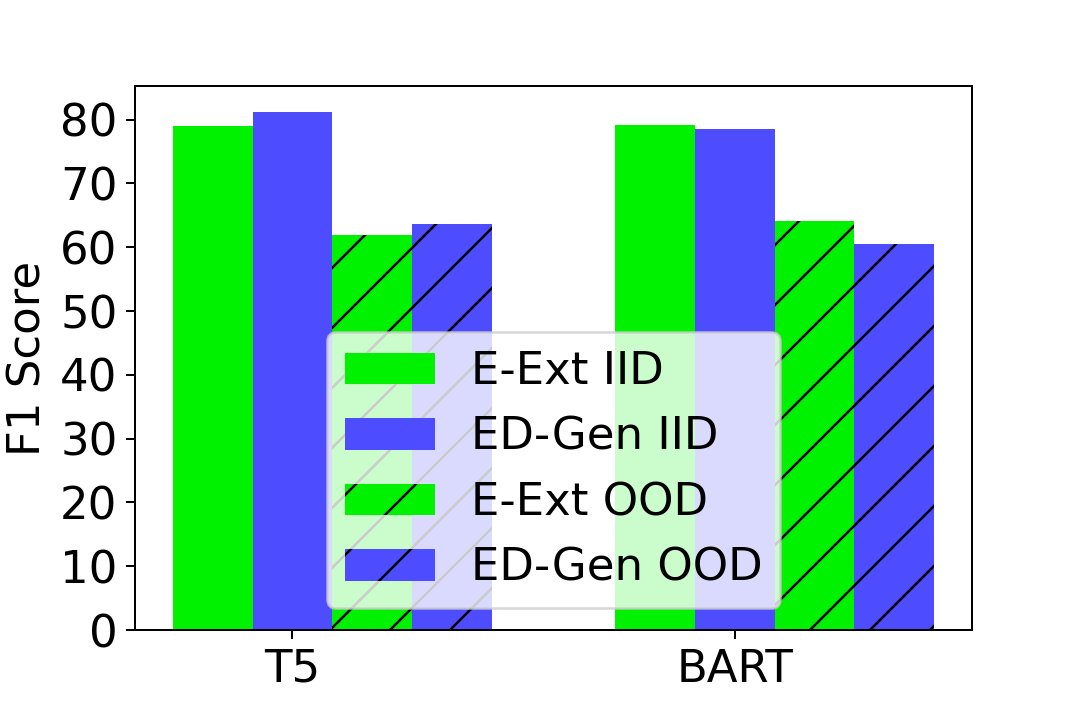}
\end{subfigure}%
\caption{
Left for single-task and right for multi-tasks settings.
For T5, ED-Gen is better than E-Ext reader; for BART, E-Ext is better than ED-Gen reader even though the former has less parameters.}
\label{fig:comp_e_ed_ext_gen}
\end{figure}

\paragraph{E-Extractive and Generative Reader} In this comparison, the extractive reader has less advantage than the generative ones since the decoder has been removed in E-Extractive reader. In Figure \ref{fig:comp_e_ed_ext_gen}, the \textcolor{black}{green} and \textcolor{black}{blue} bars compare the \textcolor{black}{E-Extractive} and \textcolor{black}{ED-Generative} reader. For T5 model, the generative reader are better than the extractive ones in both single- and multi-tasks and IID and OOD datasets. But again, this disadvantages of extractive readers might come from the smaller model size as we discussed in previous comparison. For BART model, E-Extractive reader outperforms generative reader significantly on both IID and OOD datasets and the advantage of E-Extractive reader are much more significantly in single-task learning scenario. 
To summarize, 
\begin{enumerate}[nosep,noitemsep,leftmargin=*]
\item The encoder part itself in both T5 and BART can perform well as an extractive reader. 
\item The comparison among three types of reader using BART and T5 suggests that although both PrLMs are of encoder-decoder architecture, three types of readers behave quite differently. This might caused by different pre-training objectives and knowledge. 
\item For BART model, the E-Extractive reader outperforms ED-Extractive reader and generative reader regardless of less parameters, thus should be used as an extractive reader. 
\item The BART generative reader requires large and diversified datasets to learn the QA task and thus benefits significantly from multi-task learning. 
\item For T5, the performance of generative reader consistently outperforms two types of extractive reader. The deficiency of T5-Extractive reader might be caused by less parameters.
\end{enumerate}

\subsection{Comparison within Different PrLMs} \label{sec:comp_diff_prlms}
\begin{table*}[t]
\large
\centering
\renewcommand{\arraystretch}{1.2}
{
\resizebox{\linewidth}{!}{
\begin{tabular}{@{}c|lllllll|lllllll@{}}
\toprule

\multirow{2}{*}{Model} &  \multicolumn{7}{c}{In-domain Datasets} & \multicolumn{7}{|c}{Out-of-domain Datasets}\\
\cmidrule(lr){2-8} \cmidrule(lr){9-15}
~& SQuAD & NewsQA & TQA & SQA & HQA & NQ & Avg. & DROP & RACE & BioASQ & TbQA & RE & DuoRC & Avg. \\ 
 \toprule
    \multicolumn{3}{l}{\bf Single Task Learning}\\
    \bottomrule
T5 ED-Gen  & 90.75 & 71.65 & \textbf{79.61} & \textbf{86.21} & 79.89 & 78.04 & \textbf{81.02} & 48.08 & 48.89 & 67.36 & \underline{60.30} & 84.94 & \underline{61.35} & \underline{61.82}\\
BART ED-Gen & 
78.75 & 66.20 & 67.81 & 78.89 & 73.22 & 56.58 & 70.24 & 44.22 & 43.70 & 55.59 & 45.11 & 76.83 & 55.63 & 53.51 \\
T5 E-Ext  & 92.47 & \textbf{72.63} & 76.09 & \underline{83.24} & 80.67 & \textbf{80.00} & \underline{80.85} & 53.14 & \textbf{52.06} & \textbf{71.26} & \textbf{61.92} & 85.78 & \textbf{62.80} & \textbf{64.49}\\
BART E-Ext & 
92.19 & \underline{72.20} & 73.12 & 77.19 & 80.61 & \underline{79.29} & 79.10 & 51.57 & 48.82 & \underline{68.83} & 51.29 & 86.04 & \underline{61.35} & 61.32\\
ELECTRA  & \textbf{93.39} & 60.23 & \underline{76.31} & 82.54 & \underline{80.99} & 78.78 & 78.71 & \underline{55.43} & \underline{49.80} & 66.96 & 47.80 & \underline{86.23} & 54.90 & 60.19\\
RoBERTa  & \underline{92.64} & 59.95 & 72.97 & 81.62 & \textbf{81.21} & 78.95 & 77.89 & \textbf{55.88} & 47.72 & 64.47 & 52.31 & \textbf{86.69} & 55.75 & 60.47\\

\toprule
\multicolumn{3}{l}{\bf Multi-Task Learning}\\
\bottomrule

 T5 ED-Gen  & 91.41$_{\textcolor{blue}{+0.66}}$ & 71.29$_{\textcolor{red}{-0.36}}$ & \textbf{80.01}$_{\textcolor{blue}{+0.40}}$ & \textbf{86.46}$_{\textcolor{blue}{+0.25}}$ & 79.70$_{\textcolor{red}{-0.19}}$ & 78.09$_{\textcolor{blue}{+0.05}}$ & \underline{81.16}$_{\textcolor{blue}{+0.14}}$ & 51.20$_{\textcolor{blue}{+3.12}}$ & 49.66$_{\textcolor{blue}{+0.77}}$ & 68.72$_{\textcolor{blue}{+1.36}}$ & \underline{62.90}$_{\textcolor{blue}{+2.60}}$ & 85.84$_{\textcolor{blue}{+0.90}}$ & \underline{63.76}$_{\textcolor{blue}{+2.41}}$ & 63.68$_{\textcolor{blue}{+1.86}}$\\ 
 
BART ED-Gen & 
88.63$_{\textcolor{blue}{+9.88}}$ & 68.91$_{\textcolor{blue}{+2.71}}$ & 74.91$_{\textcolor{blue}{+7.10}}$ & 82.52$_{\textcolor{blue}{+3.63}}$ & 80.53$_{\textcolor{blue}{+7.31}}$ & 75.78$_{\textcolor{blue}{+19.20}}$ & 78.55$_{\textcolor{blue}{+8.31}}$ & 55.20$_{\textcolor{blue}{+10.98}}$ & 50.04$_{\textcolor{blue}{+6.34}}$ & 63.78$_{\textcolor{blue}{+8.19}}$ & 54.81$_{\textcolor{blue}{+9.70}}$ & 80.94$_{\textcolor{blue}{+4.11}}$ & 58.47$_{\textcolor{blue}{+2.84}}$ & 60.54$_{\textcolor{blue}{+7.03}}$\\

T5 E-Ext &  92.84$_{\textcolor{blue}{+0.37}}$ & \textbf{73.51}$_{\textcolor{blue}{+0.88}}$ & \underline{77.37}$_{\textcolor{blue}{+1.28}}$ & 82.89$_{\textcolor{red}{-0.35}}$ & 81.92$_{\textcolor{blue}{+1.25}}$ & \textbf{80.74}$_{\textcolor{blue}{+0.74}}$ & \textbf{81.55}$_{\textcolor{blue}{+0.70}}$ & 59.10$_{\textcolor{blue}{+5.96}}$ & \textbf{54.01}$_{\textcolor{blue}{+1.95}}$ & \underline{71.13}$_{\textcolor{red}{-0.13}}$ & \textbf{64.90}$_{\textcolor{blue}{+2.98}}$ & 86.53$_{\textcolor{blue}{+0.75}}$ & \textbf{65.01}$_{\textcolor{blue}{+2.21}}$ & \textbf{66.78}$_{\textcolor{blue}{+2.29}}$
\\
BART E-Ext & 
92.46$_{\textcolor{blue}{+0.27}}$ & \underline{72.11}$_{\textcolor{red}{-0.09}}$ & 72.24$_{\textcolor{red}{-0.88}}$ & 76.53$_{\textcolor{red}{-0.66}}$ & 82.04$_{\textcolor{blue}{+1.43}}$ & 79.40$_{\textcolor{blue}{+0.11}}$ & 79.13$_{\textcolor{blue}{+0.03}}$ & 58.22$_{\textcolor{blue}{+6.65}}$ & 50.40$_{\textcolor{blue}{+1.58}}$ & 70.72$_{\textcolor{blue}{+1.89}}$ & 56.29$_{\textcolor{blue}{+5.00}}$ & \underline{86.79}$_{\textcolor{blue}{+0.75}}$ & 61.95$_{\textcolor{blue}{+0.60}}$ & \underline{64.06}$_{\textcolor{blue}{+2.74}}$
\\
ELECTRA & \underline{93.27}$_{\textcolor{red}{-0.12}}$ & 60.59$_{\textcolor{blue}{+0.36}}$ & 72.96$_{\textcolor{red}{-3.35}}$ & 82.03$_{\textcolor{red}{-0.51}}$ & \textbf{83.10}$_{\textcolor{blue}{+2.11}}$ & 79.16$_{\textcolor{blue}{+0.38}}$ & 78.52$_{\textcolor{red}{-0.19}}$ & \underline{62.56}$_{\textcolor{blue}{+7.13}}$ & 50.29$_{\textcolor{blue}{+0.49}}$ & \textbf{71.50}$_{\textcolor{blue}{+4.54}}$ & 54.60$_{\textcolor{blue}{+6.80}}$ & \textbf{87.14}$_{\textcolor{blue}{+0.91}}$ & 56.88$_{\textcolor{blue}{+1.98}}$ & 63.83$_{\textcolor{blue}{+3.64}}$\\

RoBERTa & 
\textbf{93.41}$_{\textcolor{blue}{+0.77}}$ & 59.56$_{\textcolor{red}{-0.39}}$ & 72.23$_{\textcolor{red}{-0.74}}$ & 80.98$_{\textcolor{red}{-0.64}}$ & 82.37$_{\textcolor{blue}{+1.16}}$ & \underline{79.55}$_{\textcolor{blue}{+0.60}}$ & 78.02$_{\textcolor{blue}{+0.13}}$ & \textbf{64.47}$_{\textcolor{blue}{+8.59}}$ & \underline{51.81}$_{\textcolor{blue}{+4.09}}$ & 69.15$_{\textcolor{blue}{+4.68}}$ & 53.68$_{\textcolor{blue}{+1.37}}$ & 86.31$_{\textcolor{red}{-0.38}}$ & 56.06$_{\textcolor{blue}{+0.31}}$ & 63.58$_{\textcolor{blue}{+3.11}}$\\

\bottomrule 
\end{tabular}
}
}
\caption{Comparison of readers based on the different PrLMs by F1 Score. Inference length of T5 is full length of context, 512 for ELECTRA, and 1024 for BART and RoBERTa.  TQA: TriviaQA; SQA:SearchQA; HQA:HotpotQA; NQ: NaturalQuestions; TbQA:TextbookQA; RE:RelationExtraction. \textbf{Bold} numbers denote for the best result and \underline{underline} numbers for the second best.}
\label{tab:compare_diff_prlm}
\end{table*}

The previous section compares the generative and extractive readers using the same PrLMs and both PrLMs are encoder-decoder models. On one hand, such comparison reduces the impacts of PrLMs architecture and pre-owned knowledge. On the other hand, it raises two concerns. First, whether extractive readers using an encoder-decoder PrLMs are good for representatives of extractive readers? After all, encoder-only PrLMs are more standard choice for extractive readers in most previous work. Second, whether the smaller size of the extractive reader cause its deficiency compared to the generative one, particularly that the T5 E-Extractive reader is half size of the T5 generative reader in previous comparison. 

To clear out the first concern, here, we present the comparison cross different PrLMs including standard encoder-only models for extractive readers.
To address the second concern, we carefully select the model size so that each model is of relative comparable size. 

\paragraph{The Selection of Each Model's size}
We use the encoder in T5 large model for the T5 E-Extractive reader so that it is of similar size as RoBERTa and ELECTRA extractive readers ($\sim$330M)\footnote{ Note that the T5 PrLM is already trained on SQuAD, while others do not. However, based on the results on SQuAD, T5 does not have advantage over other models on this dataset.}.
When using BART PrLMs for extractive reader, we only use BART E-Extractive reader but not ED-Extractive reader because the former performs better even though it has less parameters (204M) than the later one has larger size. 
T5 generative reader is also smaller (223M), but this is better than using T5 large generative reader to compare with others, which is way too larger than other readers (737M). 
For BART generative reader, it is larger than other readers (406M). 
One potential issue for the abovementioned setting is that even though we choose the best comparison setting, still each model size are different, and thus if a model perform inferior than others, it might due to the smaller model size. 
However, the following conclusion we draw does not effect by this issue.

\paragraph{Are Encoder-decoder PrLMs Good for Extractive Readers?} Based on Table \ref{tab:compare_diff_prlm}, we find that encoder-decoder PrLMs outperform encoder-only PrLMs as extractive readers on average. 
Both T5 and BART E-Extractive readers perform better than RoBERTa and ELECTRA on IID and OOD datasets under single- as well as multi-task learning regardless of less parameters of T5 and BART. 
This observation is exciting since instead of using standard encoder-only PrLMs for extractive reader, encoder-decoder PrLMs are actually better choice. 

\paragraph{Which reader generalize better on OOD?} 
The extractive reader generalize better on OOD datasets. 
In both single- and multi-task learning, T5 E-Extractive reader shows the best performance, especially beating the BART generative reader even though the latter one has more parameters. BART E-Extractive reader also generalize well on OOD, and it also beats the BART generative reader even though the former has less parameters than the later. 

\paragraph{Which PrLM is the best?} Based on Table \ref{tab:compare_diff_prlm}, we see that T5 is the best among four PrLMs in both single- and multi-tasks learning scenario on IID as well as OOD datasets. We observe two advantages of T5 over other PrLMs. First, T5 is much better than ELECTRA and RoBERTa on NewsQA data. In both single- and multi- task learning, RoBERTa and ELECTRA achieve around $60\%$ F1 score on NewsQA, while both T5 extractive and generative reader achieved higher than $70\%$ F1 score, yielding more than $10\%$ improvements. Second, T5 is better at long context dataset. In IID, TQA and SQA, T5 ED-Generative reader outperforms other readers at least $3.30\%$ and $3.67\%$ in single-task,  $7.05\%$ and $4.43\%$ in multi-task learning. On OOD datasets, TbQA and DuoRC, T5 E-Extractive reader is better than others at least by $9.61\%$ and $1.45\%$ in single-task, $8.61\%$ and $3.06\%$ in multi-task. We would like to mention that this advantage of T5 is conditioned on using full inference length, when using short input length such as 512, this advantage does not exhibit as we shown in \S\ref{sec:length_effect}.

\paragraph{Which PrLM benefits more from Multi-task Learning?}
While multi-task learning is in general beneficial for all PrLMs, we find BART benefits the most from multi-task learning, especially for the generative reader. For example, on IID datasets. BART generative reader improves more than $8\%$ on average while all other readers improves less than $1\%$. Similarly for OOD datasets, the improvement of multi-task learning on BART generative reader are more significant than other readers. 
To summarize,
\begin{enumerate}[nosep,noitemsep,leftmargin=*]
\item Encoder-decoder PrLMs are in fact can be used as extractive readers, they are even better than the conventional choice (encoder-only PrLMs) of extractive readers on average. 
\item Extractive readers performs better than the generative readers on OOD datasets, especially for the extractive readers based on the encoder-decoder PrLMs. 
\item T5 is the best among four PrLMs for it performs better in News domain and the long context. And the advantage of T5 is conditioned on using full inference length. 
\item While in general multi-task learning benefits for all PrLMs, BART PrLM benefits the most. 
\end{enumerate}

\subsection{In-Depth Diagnosis} \label{sec:additional_findings}
We investigate the behavior of extractive and generative models in long and short context and predicting answers which include rare characters. 
Multi-task models in \S\ref{sec:comp_diff_prlms} are chosen for comparison. 
\subsubsection{Long and Short Context} \label{sec:long_short_length}
\begin{figure*}
    \centering
    \begin{minipage}{0.45\textwidth}
        \centering
        \includegraphics[width=0.9\textwidth]{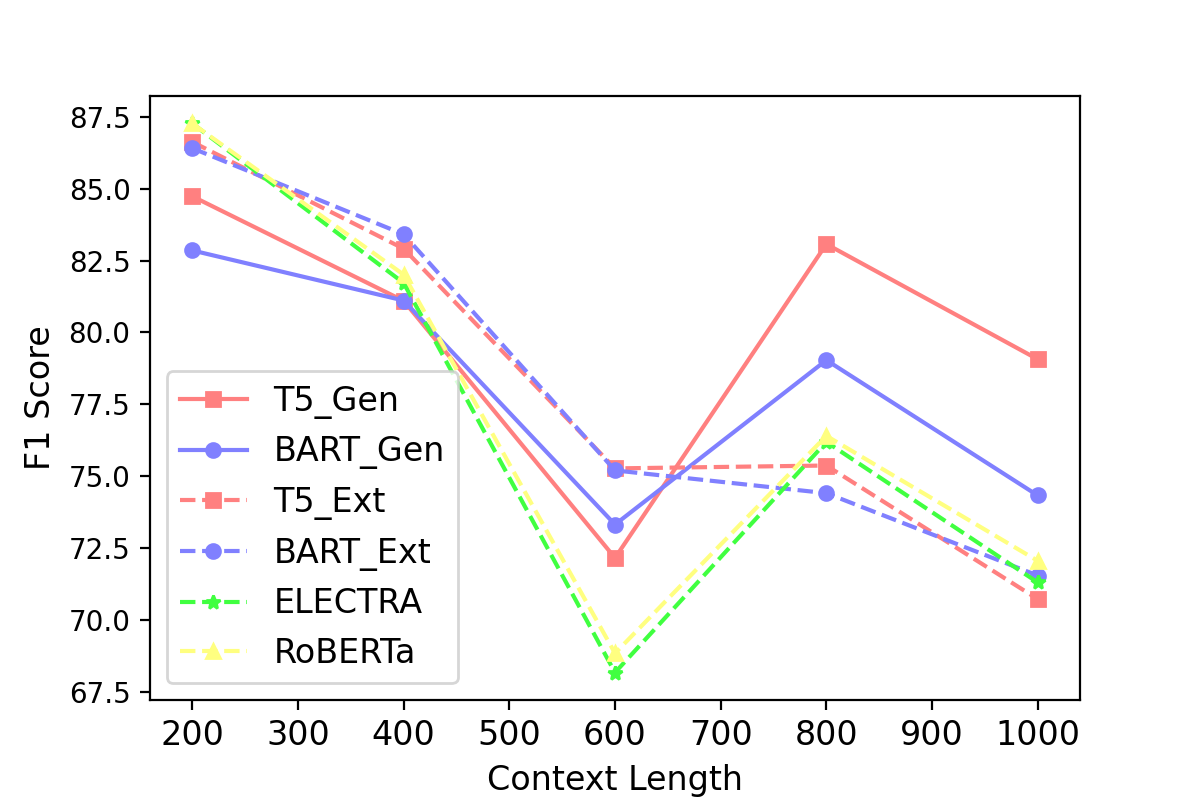} 
        
    \end{minipage}\hfill
    \begin{minipage}{0.45\textwidth}
        \centering
        \includegraphics[width=0.9\textwidth]{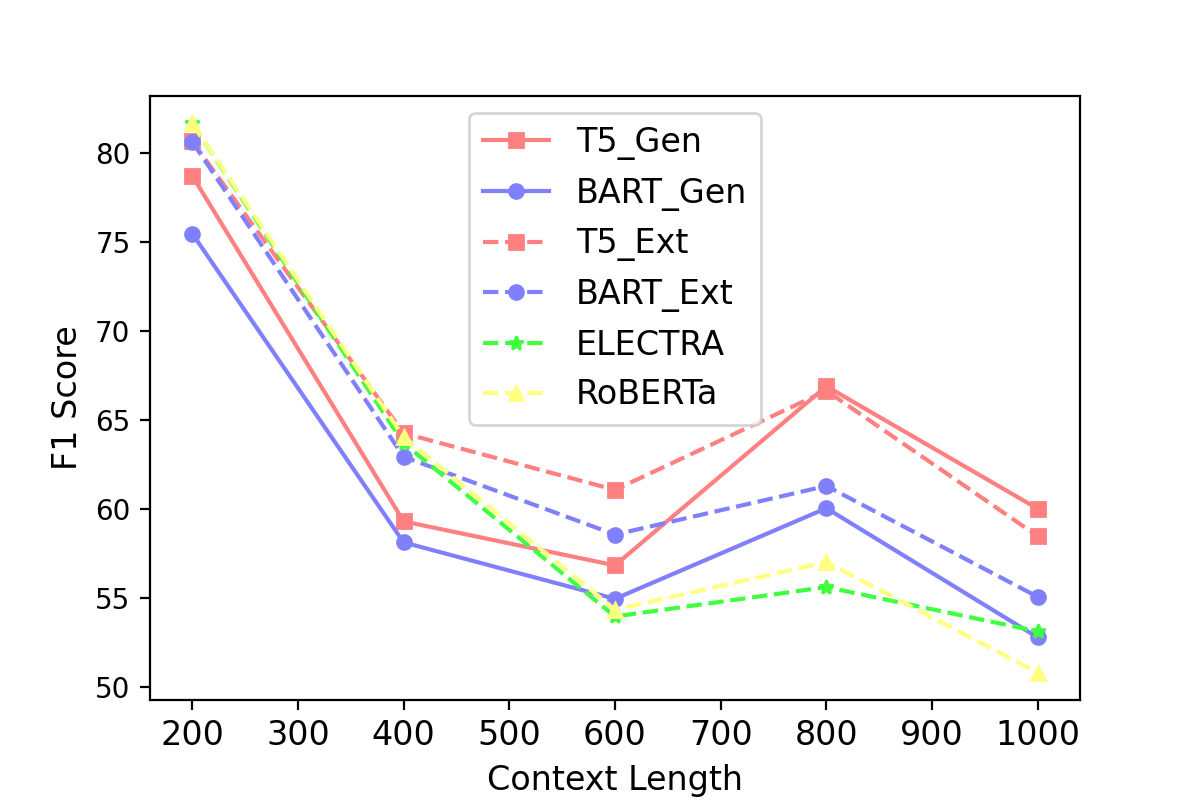} 
    \end{minipage}
    \caption{Comparison among generative and extractive readers on different length of the question and context. Left part for IID and right part for OOD datasets. Dash line for extractive and solid line for generative readers.}
    \label{fig:compare_length}
\end{figure*}

As we discussed in previous section that generative readers have advantage over extractive counterparts. To further support this trend, we divide the testing sets into five subsets, where we count the total words in question and context, and choose five thresholds, 2/4/6/8/10 hundreds. 
It is worth to mention that since all extractive readers use the window-stride strategy
(i.e. if the input length is longer than the maximum length, then the input is segmented into multiple inputs),
so that the entire context is observable for extractive readers. 

From Figure \ref{fig:compare_length}, we have two observations. First, on IID datasets, for questions and contexts with less than 600 words, the extractive ones always perform better than the generative ones (the dash lines are higher than the solid ones), but when the length are more than 600 words, the generative ones consistently outperform the extractive ones. This suggests that the extractive readers performs better in the short context while the generative readers perform better in long context. Second, on OOD datasets, T5 generative reader still presents advantage in the long context (more than 600 words), while BART generative reader performs worse than the extractive one in both short and long context. But the gap between the BART generative and extractive readers is less on the long context compared to the short context. It might suggest that the extractive reader has better generalization capacity than the generative one thus the advantage of generative reader in long context is weakened. 

\subsubsection{Rare Characters in Answer} 
We find that some answers of testing sets include rare characters such as \textit{\'{n}} and \textit{ł} (119 are found), thus we divide the testing sets into two subsets, one is the normal answer set where the answer does not have rare characters\footnote{Rare characters are any characters which are not belongs to the printable characters in the string library of Python. The printable characters include lower and upper case alphabets, digits, punctuation, and white-space.}, the other one is with rare characters. The percentage of rare cases for IID and OOD datasets is 1.4\% and 2\%, respectively. 

From Table \ref{tab:rare}, we have two observations. First, in normal case, the performance of extractive and generative readers are relatively comparable on both IID and OOD datasets, but in rare case, the extractive readers are better than the generative ones
This suggests that the extractive reader has better generalization than the generative ones. Second, we see that the rare tokens has worse impact on T5 than BART generative readers in both in- and out-of-domain datasets. 
Further investigation finds that 94 out of 119 
rare characters can not be represented by T5 tokenizer (i.e. T5 tokenizer uses `<unk>' special tokens to represent these characters), and tends to ignore these special characters in the generation time as the two examples shown in Table \ref{tab:rare_example}. Differently, BART tokenizer can represent all rare characters. 
Improving generative readers performance in predicting rare answers is an important future work. 
To summarize, 
\begin{enumerate}[nosep,noitemsep,leftmargin=*]
\item Extractive readers performs better than the generative reader on short context, but generative one performs better on long context. 
\item Generative readers performs worse in predicting answers with rare characters, and T5 performs worse than BART.
\end{enumerate}

        
\begin{table}[t]
\centering
 \resizebox{0.98\linewidth}{!}{
\begin{tabular}{c|c|c|c|c|c|c|c}
    \toprule
    \multirow{2}{*}{\bf Answer  type} & \multirow{2}{*}{\bf Domain}  & \multicolumn{2}{c}{\bf Gen } & \multicolumn{4}{|c}{\bf Ext } \\
    \cmidrule(lr){3-4}\cmidrule(lr){5-8}
     ~& ~ & T5 & BART & T5 & BART & Ro &  EL\\
     \hline
        \multirow{2}{*}{Rare}
        & IID   & 68.97 & 73.64  & 77.79 & \textbf{78.54} & 78.64 & 78.18\\
        & OOD   & 59.25 & 79.84  & \textbf{85.22} & 84.95 & 80.73 & 86.94\\
        \midrule
        \multirow{2}{*}{Normal}
        & IID   & \textbf{82.71} & 80.02  & 79.98 & 79.95 & 80.35 & 78.18\\
        & OOD   & 68.28 & 64.19  & \textbf{69.9} & 66.91 & 67.75 & 68.12\\
        
    \toprule
    \end{tabular}
    }
\caption{Compare extractive and generative readers in terms of rare and normal answers. Ro for RoBERTa and EL for ELECTRA.}
\label{tab:rare}
\end{table}
\begin{table}[t]
    \centering 
    \small
\resizebox{0.92\linewidth}{!}{
    \begin{tabular}{@{}p{0.35\linewidth}p{0.25\linewidth}p{0.25\linewidth}@{}}
        \toprule
        \textbf{Question} & \textbf{Answer} & \textbf{Prediction} \\
        \toprule
        Who was one of the most famous people born in Warsaw? 
        & Maria Sk\textcolor{blue}{ł}odowskacurie
        & Maria Skodowska-Curie\\
        
    \midrule
    What museum preserves the memory of the crime?
        & Katy\textcolor{blue}{ń} Museum
        & Katy Museum \\
        
    \bottomrule
    \end{tabular}
    }
    \caption{Examples of questions with answers containing rare characters and the prediction of T5-Gen.}
    \label{tab:rare_example}
\end{table}

\section{Conclusion and Future Work}
We systematically compare the extractive and generative readers for QA tasks. 
Two sets of experiments are designed to control the effects of different PrLMs and the size of models. 
By conducting experiments on 12  QA datasets, 
our findings provide guidelines on how to choose extractive or generative readers given their strength and weakness. 
Investigating the reason behind the observations and improving the generative and extractive reader will be interesting research questions for future. 


\newpage
\bibliography{anthology,custom}
\bibliographystyle{acl_natbib}
\newpage
\appendix
\section{More Details of MRQA Datasets}
\label{apd:datasets}
\begin{figure*}[h]
\centering
\resizebox{\linewidth}{!}{
\begin{subfigure}{0.3\textwidth}
  \centering
  \includegraphics[width=\linewidth]{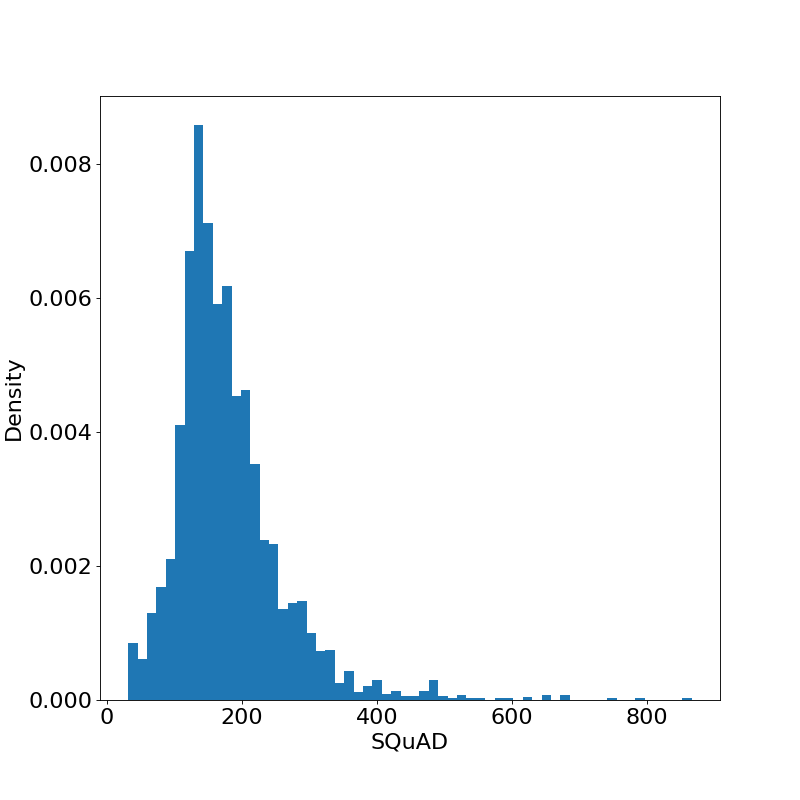}
  
  \label{fig:sub1}
\end{subfigure}%
\begin{subfigure}{0.3\textwidth}
  \centering
  \includegraphics[width=\linewidth]{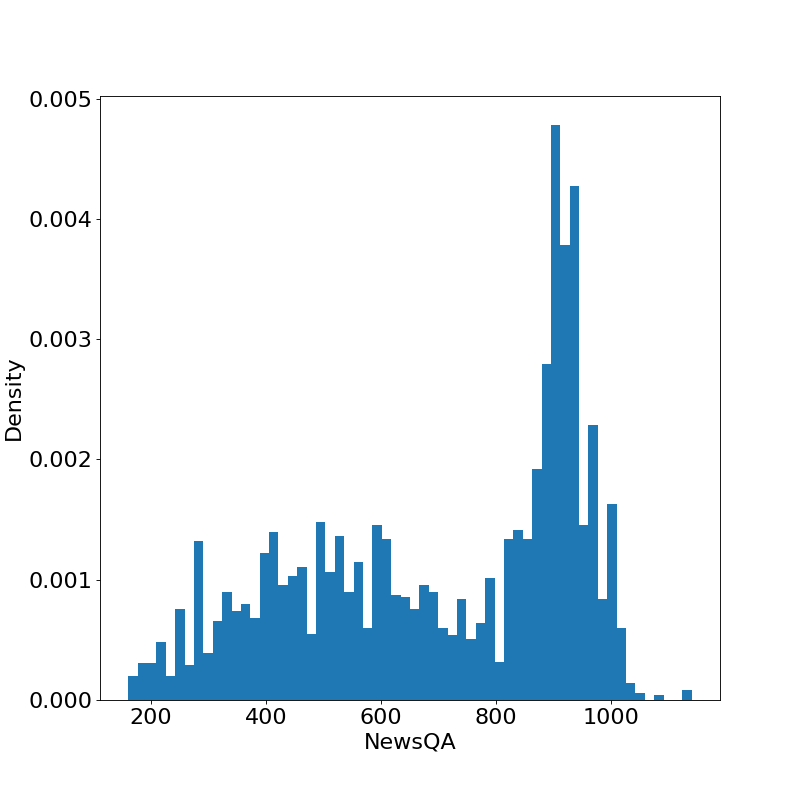}
  
  \label{fig:sub2}
\end{subfigure}%
\begin{subfigure}{0.3\textwidth}
  \centering
  \includegraphics[width=\linewidth]{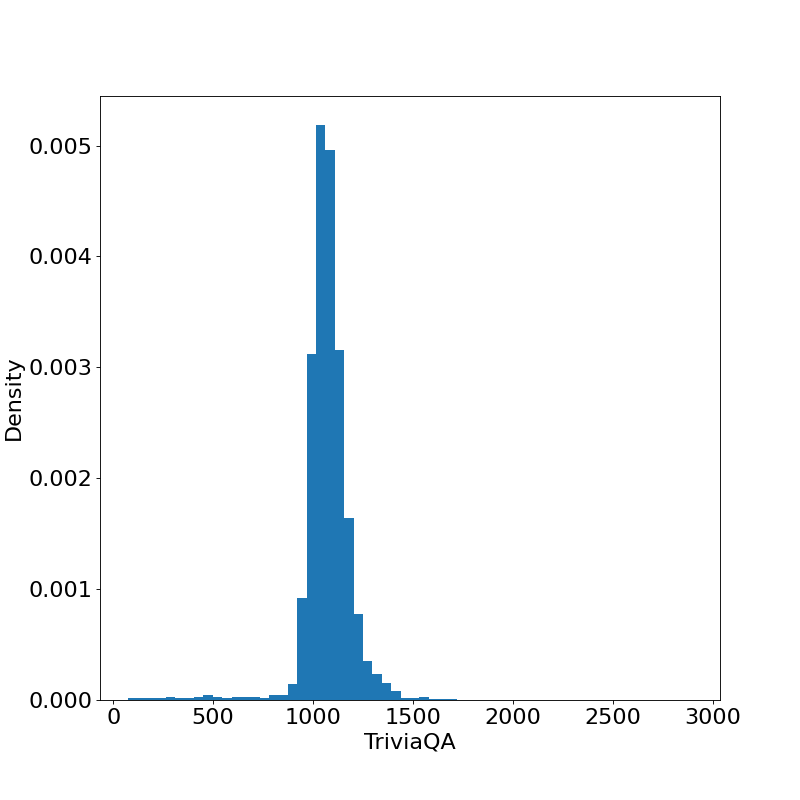}
  
  \label{fig:sub3}
\end{subfigure}

}
\resizebox{\linewidth}{!}{
\begin{subfigure}{0.3\textwidth}
  \centering
  \includegraphics[width=\linewidth]{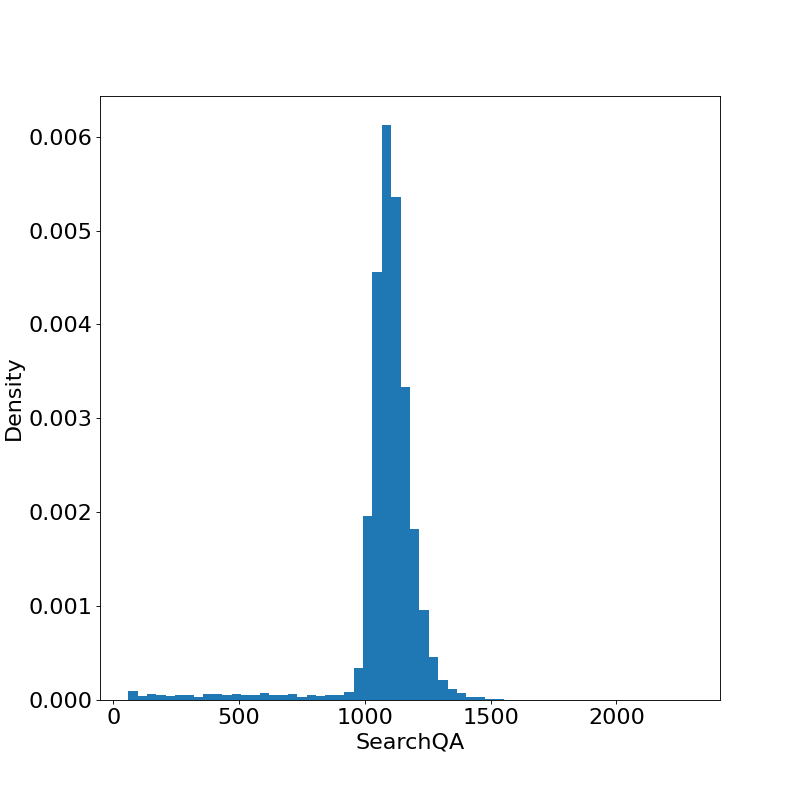}
  
  \label{fig:sub1}
\end{subfigure}%
\begin{subfigure}{0.3\textwidth}
  \centering
  \includegraphics[width=\linewidth]{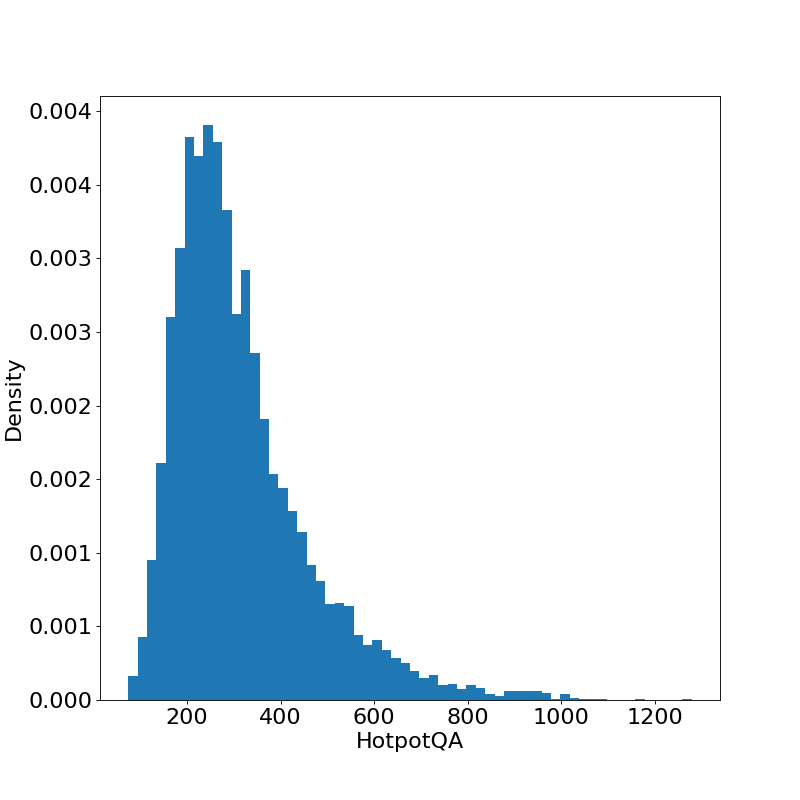}
  
  \label{fig:sub2}
\end{subfigure}%
\begin{subfigure}{0.3\textwidth}
  \centering
  \includegraphics[width=\linewidth]{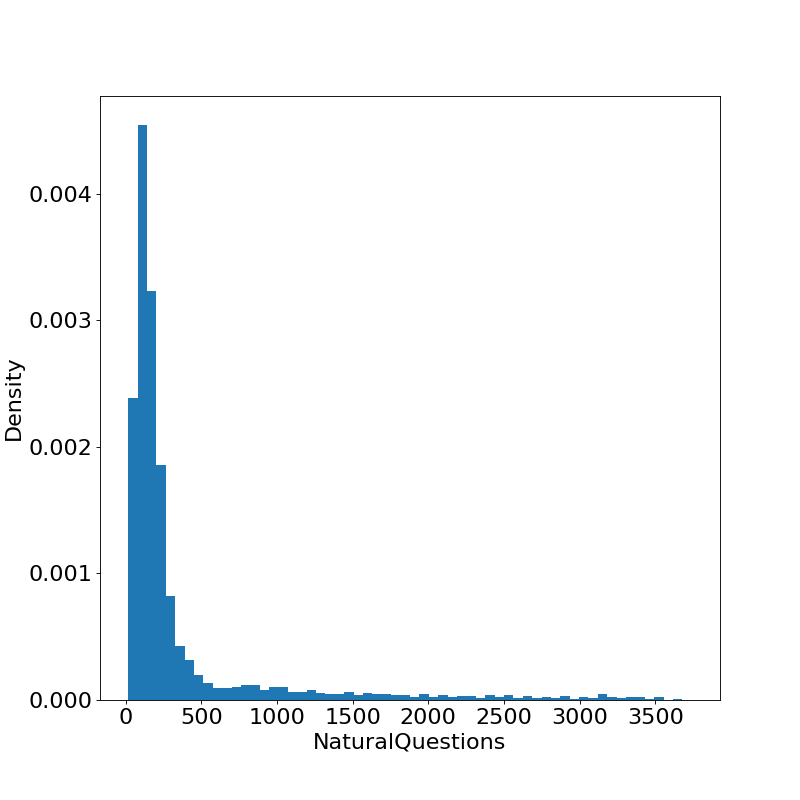}
  \label{fig:sub3}
\end{subfigure}
}
\caption{Context Length Histogram of In-domain dataset}
\label{fig:histogram-in}
\end{figure*}

\begin{figure*}[h]
\centering
\resizebox{\linewidth}{!}{
\begin{subfigure}{0.3\textwidth}
  \centering
  \includegraphics[width=\linewidth]{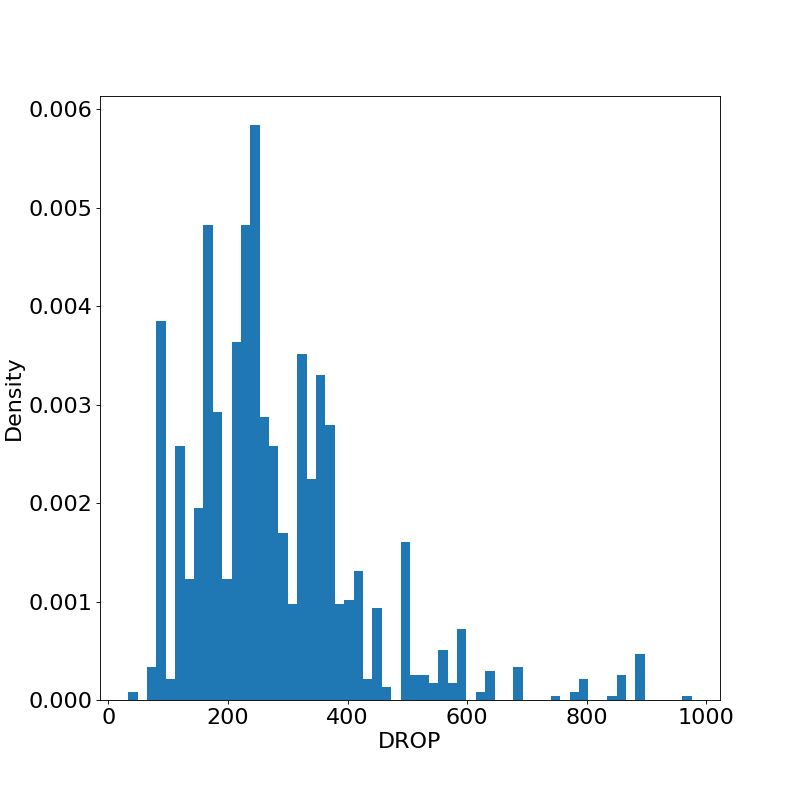}
  
  \label{fig:sub1}
\end{subfigure}%
\begin{subfigure}{0.3\textwidth}
  \centering
  \includegraphics[width=\linewidth]{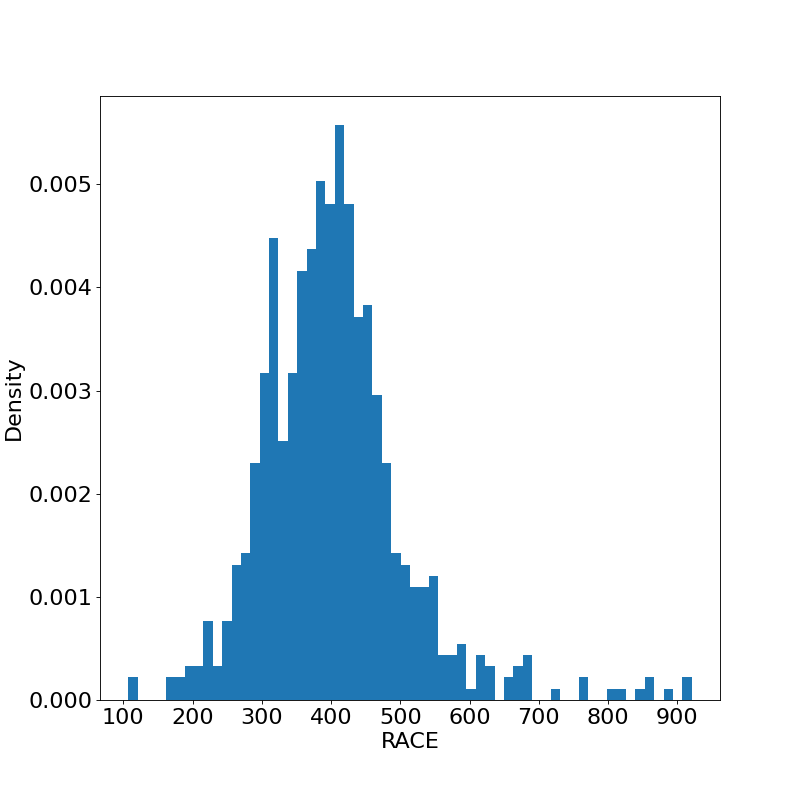}
  
  \label{fig:sub2}
\end{subfigure}%
\begin{subfigure}{0.3\textwidth}
  \centering
  \includegraphics[width=\linewidth]{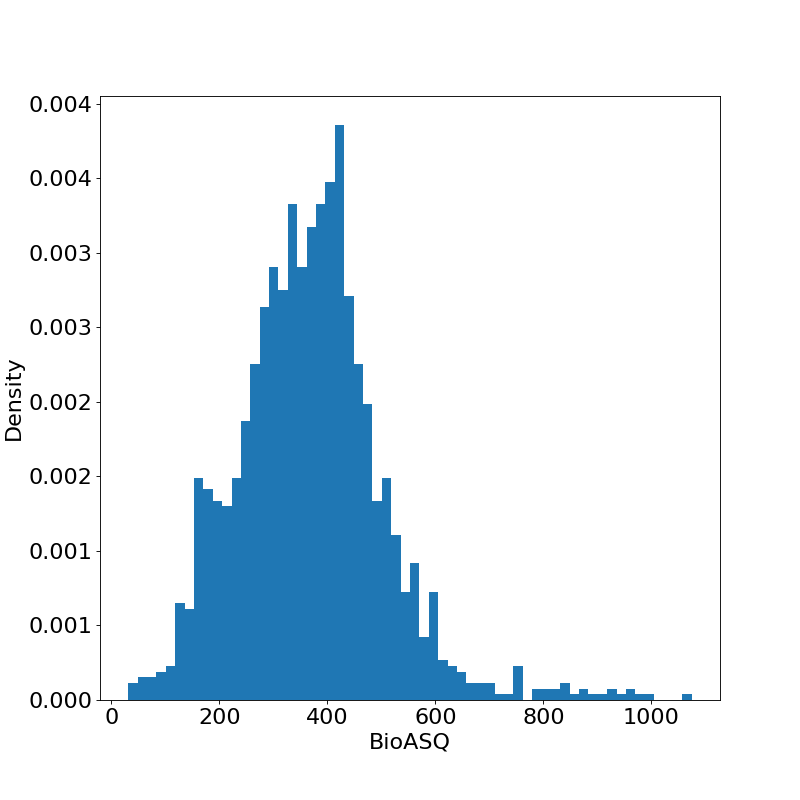}
  
  \label{fig:sub3}
\end{subfigure}

}
\resizebox{\linewidth}{!}{
\begin{subfigure}{0.3\textwidth}
  \centering
  \includegraphics[width=\linewidth]{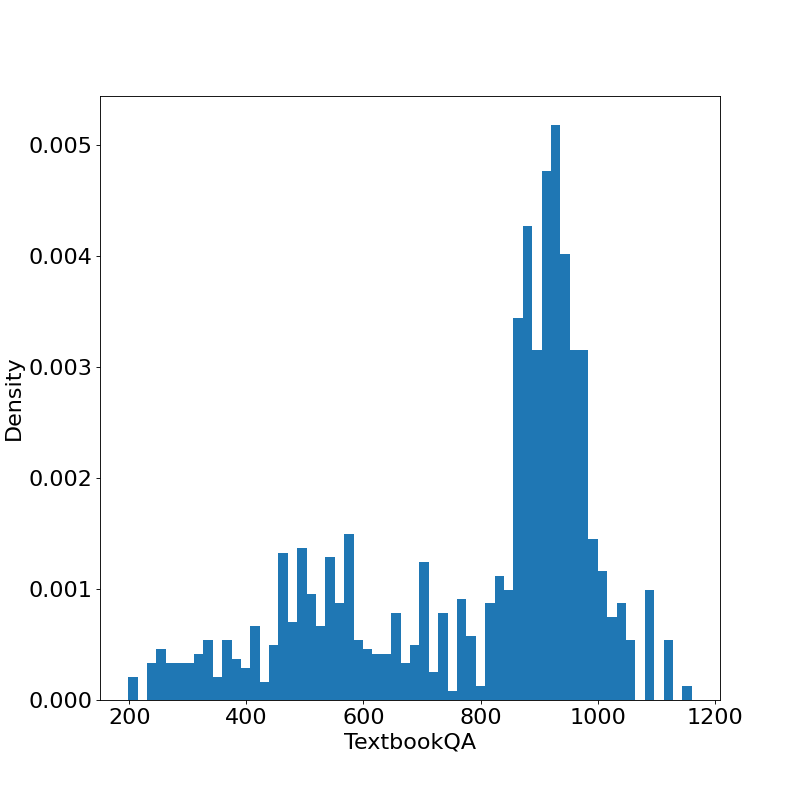}
  
  \label{fig:sub1}
\end{subfigure}%
\begin{subfigure}{0.3\textwidth}
  \centering
  \includegraphics[width=\linewidth]{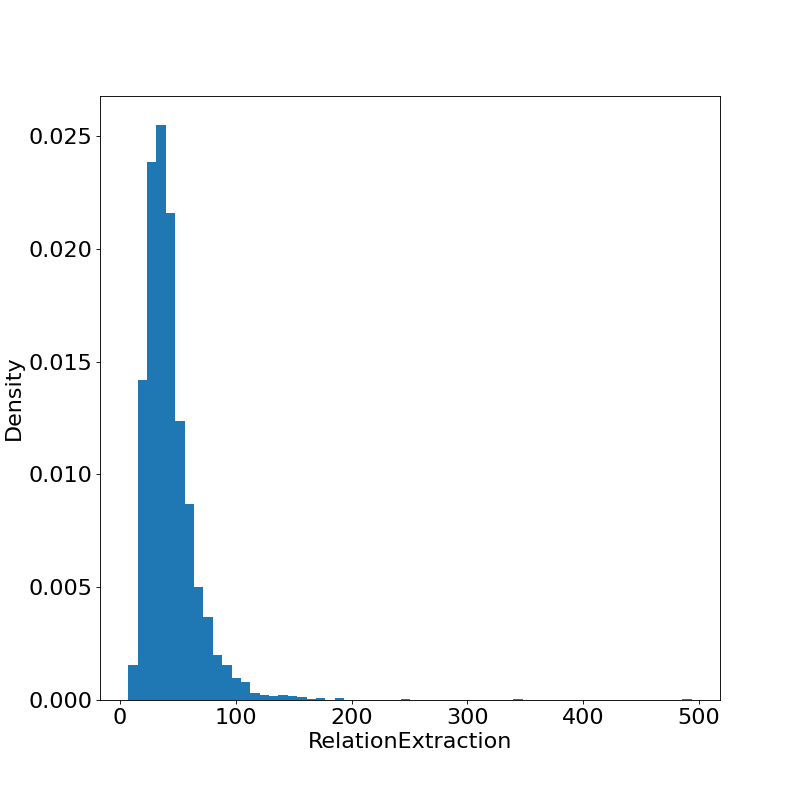}
  
  \label{fig:sub2}
\end{subfigure}%
\begin{subfigure}{0.3\textwidth}
  \centering
  \includegraphics[width=\linewidth]{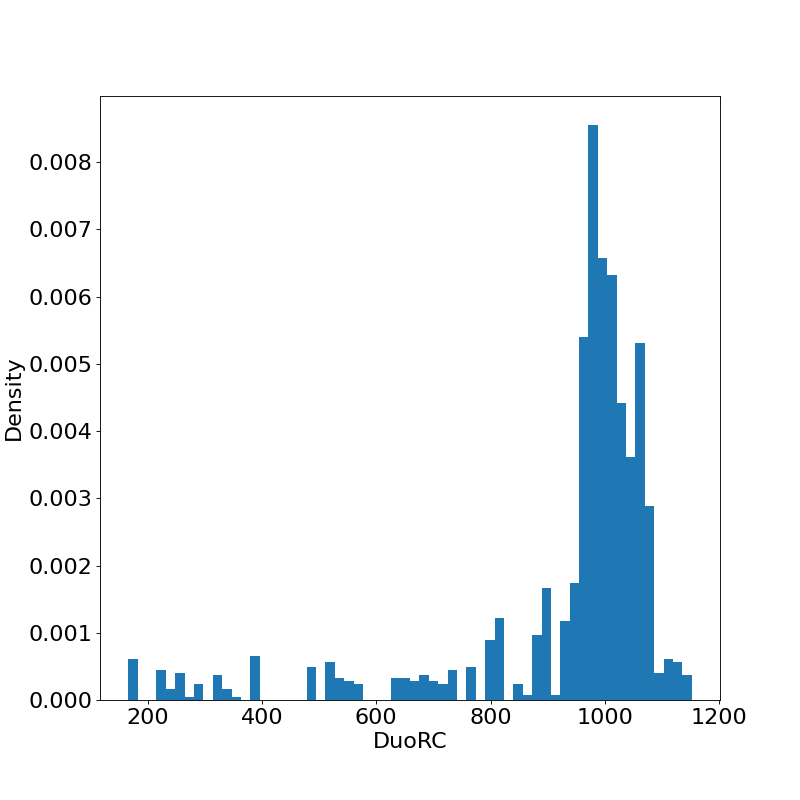}
  \label{fig:sub3}
\end{subfigure}
}
\caption{Context Length Histogram of out-domain dataset}
\label{fig:histogram-out}
\end{figure*}

MRQA provides six datasets for training and six for out-of-domain evaluations. In Table \ref{tab:dataset_source}, we present the source of each datasets, and we can see that the domains are diversified. Figure \ref{fig:histogram-in} and \ref{fig:histogram-out} show the histogram of the context length of IID and OOD dataset. The distribution shows that some datasets are mainly short, some are mainly long, and others are the combination of short and long. We use short annotation for some datasets, TQA: TriviaQA; SQA:SearchQA; HQA:HotpotQA; NQ: NaturalQuestions; TbQA:TextbookQA; RE:RelationExtraction. 
\begin{table}[h]
\centering
\small

 \resizebox{0.97\linewidth}{!}{
\begin{tabular}{@{}p{0.2\linewidth}p{0.75\linewidth}@{}}
    \toprule
     Dataset  & Source \\
     \hline
     SQuAD & Wikipedia\\
     NewsQA & News article \\
     TQA & Trivia and quiz-league websites  \\
     SQA & Jeopardy! TV show  \\
     HQA & Wikipedia\\
     NQ & Wikipedia \\
     DROP & Wikipedia \\
     RACE & English reading comprehension exams for middle and high school
\\
     BioASQ & Science (PubMed) articles \\
     TbQA &  Lessons from middle school Life Science, Earth Science, and Physical Science textbooks
\\
     RE & Wikiread \\
     DuoRC & wikipedia \\
    \toprule
    \end{tabular}
    }
\caption{The source of each dataset}
\label{tab:dataset_source}
\end{table}

\section{Training Setup} \label{apd:setup}

We use Huggingface~\cite{wolf-etal-2020-transformers} implementation and Pytorch~\cite{NEURIPS2019_9015} to train each model. All model are trained on 4 GTX1080 GPUs in 4 epochs with a learning rate of 1e-4, batch size of 128, random seed 1234.
While we fix these hyperparameters for all models, we get similar results as the original paper (i.e. the difference in terms of F1 are mostly within 2 percent.) 
In details, on SQuAD dataset, RoBERTa in \cite{Liu2019RoBERTaAR} and in ours achieves 94.6 and 92.64 F1 scores, respectively;  BART in \cite{lewis2020bart} and in ours achieves 94.6 and 92.51 F1 scores, respectively; ELECTRA in \cite{Clark2020ELECTRAPT} and in ours achieves 94.2 and 93.39 F1 scores, respectively; T5 in \cite{Raffel2020ExploringTL} and in ours achieves 80.88 and 82.56 EM scores, respectively. 

\section{Two Input Format}\label{apd:two_input_format}

When fine-tuning generative reader on question answering task, some special words are added before the real input to denote the type of task. In an extractive reader, usually, there are no special words added. Here, we evaluate these two formats for T5 and BART generative reader. Particularly, given a question {\bf Q} and a context {\bf C}, format 1 is to add the ``question:'' and ``context:'' in front of the real question and context such that the input is \{{\textit{question:} \bf Q} [SEP] \textit{context: }{\bf C}\}; and format 2 is without these special words such that the input is  \{{\bf Q} [SEP] {\bf C}\}. To keep the training process be efficient, we evaluate on two datasets SearchQA and HotpotQA, instead of all datasets. Table \ref{tab:two_input_format} shows that format 1 yields slightly better performance for T5 and much better performance for BART on SQA datasets, and thus we use this format for all generative reader.

\begin{table}[h]
\centering
\renewcommand{\arraystretch}{1.2}
{
\resizebox{0.9\linewidth}{!}{
\small
\begin{tabular}{c|c|cc|cc}
\toprule
 \multirow{2}{*}{Model} & \multirow{2}{*}{Format} & \multicolumn{2}{c}{SQA} & \multicolumn{2}{|c}{HQA}\\
\cmidrule(lr){3-4} \cmidrule(lr){5-6}
~& ~ & EM & F1 & EM & F1 \\ 
 \toprule
\multirow{2}{*}{T5} & 1 & 81.07 & 86.21  & 64.04 & 79.89\\ 
~ & 2 & 80.65 & 85.76  & 63.23 & 79.42\\
\midrule
\multirow{2}{*}{BART} & 1  & 72.86 & 78.89  & 55.77 & 73.22\\
~ & 2 & 49.28 & 58.00  & 55.72 & 73.20  \\
\bottomrule
\end{tabular}
}
}
\caption{Comparison between different input format on two datasets. Format1 means input with ``question:'' and ``context:'' as format1, and format2 means without.}
\label{tab:two_input_format}
\end{table}

\section{Answer Length of Generative Reader} \label{apd:gen_ans_length}
For the generative reader, we tried different maximum lengths of the generated answer: 16, 32, and 64. Table \ref{tab:gen_answer_len} shows that increasing the length of the target does not make improvement, this might be because the answer in the testing data is usually short and thus length of 16 is sufficient. 

\begin{table*}[t]
\centering
\renewcommand{\arraystretch}{1.2}
{
\resizebox{\linewidth}{!}{
\begin{tabular}{c|ccccccc|ccccccc}
\toprule

\multirow{2}{*}{Length} &  \multicolumn{7}{c}{IID Datasets} & \multicolumn{7}{|c}{OOD Datasets}\\
\cmidrule(lr){2-8} \cmidrule(lr){9-15}
~& SQuAD & NewsQA & TQA & SQA & HQA & NQ & Avg. & DROP & RACE & BioASQ & TbQA & RE & DuoRC & Avg. \\ 
 \toprule
 16  &91.41 & 71.29 & 80.01 & 86.46 & 79.7 & 78.09 & 81.16 & 51.2 & 49.66 & 68.72 & 62.9 & 85.84 & 63.76 & 63.68\\ 
32 & 91.41 & 71.29 & 80.01 & 86.46 & 79.7 & 78.09 & 81.16 & 51.2 & 49.66 & 68.72 & 62.9 & 85.84 & 63.76 & 63.68\\
64 & 91.41 & 71.29 & 80.01 & 86.46 & 79.7 & 78.09 & 81.16 & 51.2 & 49.66 & 68.72 & 62.9 & 85.84 & 63.76 & 63.68\\
\midrule 
16 & 88.63 & 68.91 & 74.91 & 82.52 & 80.53 & 75.78 & 78.55 & 55.2 & 50.04 & 63.78 & 54.81 & 80.94 & 58.47 & 60.54\\
32 & 88.72 & 69.05 & 74.91 & 82.52 & 80.56 & 75.93 & 78.61 & 55.21 & 50.05 & 63.74 & 54.82 & 80.92 & 58.49 & 60.54\\
64 & 88.72 & 69.05 & 74.91 & 82.52 & 80.56 & 75.93 & 78.61 & 55.21 & 50.05 & 63.74 & 54.82 & 80.92 & 58.49 & 60.54\\
\bottomrule
\end{tabular}
}
}
\caption{Performance of using different Answer length for generative reader. First block is the result for T5 model and the second block for BART model.}
\label{tab:gen_answer_len}
\end{table*}

\section {Inference Length} \label{apd:infer_length}
We present the results of using 512 and 1024 length and full length in Table \ref{tab:table_512}, \ref{tab:table_1024}, \ref{tab:table_max} separately. Note that due the tokenization approach adapted by each model, for Electra using 1024 or full length is same as using 512, for RoBERTa and BART, using full length is the same as length 1024. Furthermore, the detailed performance of each single task model is given in Table \ref{tab:apd_table_f1_best}, using the best inference of each model, i.e. full length for T5, 1024 for RoBERTa and BART, and 512 for ELECTRA.   
\begin{table*}[t]
\large
\centering
\renewcommand{\arraystretch}{1.2}
{
\resizebox{\linewidth}{!}{
\begin{tabular}{c|lllllll|lllllll}
\toprule

\multirow{2}{*}{Model} &  \multicolumn{7}{c}{In-domain Datasets} & \multicolumn{7}{|c}{Out-domain Datasets}\\
\cmidrule(lr){2-8} \cmidrule(lr){9-15}
~& SQuAD & NewsQA & TQA & SQA & HQA & NQ & Avg. & DROP & RACE & BioASQ & TbQA & RE & DuoRC & Avg. \\ 
 \toprule
    \multicolumn{3}{l}{\bf Single Task Learning}\\
    \bottomrule
T5 E-Ext (B)  & 90.12 & 59.38 & 67.39 & 77.14 & 76.95 & 75.56 & 74.42 & 41.17 & 45.46 & 64.92 & 46.69 & 84.48 & 52.61 & 55.89 \\
T5 E-Ext (L)  & 92.39 & 59.62 & 70.22 & 78.52 & 80.06 & 77.93 & 76.46 & 52.73 & 51.38 & 69.99 & 49.76 & 85.78 & 54.82 & 60.74\\
T5 ED-Ext (B) & 90.57 & 58.00 & 66.87 & 77.66 & 78.68 & 76.69 & 74.75 & 45.49 & 45.56 & 66.99 & 48.66 & 84.91 & 51.03 & 57.11\\
T5 ED-Gen (B) & 90.63 & 66.74 & 73.45 & 82.75 & 78.81 & 75.10 & 77.91 & 48.07 & 47.54 & 67.33 & 46.19 & 84.94 & 43.49 & 56.26\\
BART E-Ext (L) & 92.15 & 62.31 & 72.84 & 79.99 & 80.52 & 78.86 & 77.78 & 50.91 & 48.83 & 68.18 & 47.19 & 86.04 & 56.89 & 59.67\\
BART ED-Ext (L) & 92.50 & 58.81 & 72.11 & 80.33 & 80.30 & 78.57 & 77.10 & 54.74 & 47.13 & 66.05 & 47.00 & 86.15 & 54.66 & 59.29\\
BART ED-Gen (L) & 78.72 & 63.18 & 69.22 & 79.39 & 72.72 & 56.09 & 69.89 & 44.04 & 43.64 & 53.79 & 38.44 & 72.17 & 45.84 & 49.65\\

ELECTRA (L) & 93.39 & 60.23 & 76.31 & 82.54 & 80.99 & 78.78 & 78.71 & 55.43 & 49.80 & 66.96 & 47.80 & 86.23 & 54.90 & 60.19\\
RoBERTa (L) & 92.67 & 59.32 & 72.52 & 81.34 & 80.88 & 78.82 & 77.59 & 55.02 & 48.18 & 64.66 & 52.42 & 86.65 & 54.98 & 60.32\\
\toprule
\multicolumn{3}{l}{\bf Multi-Task Learning}\\
\bottomrule
T5 E-Ext (B)  & 90.76$_{\textcolor{blue}{+0.64}}$ & 61.69$_{\textcolor{blue}{+2.31}}$ & 68.95$_{\textcolor{blue}{+1.56}}$ & 77.58$_{\textcolor{blue}{+0.44}}$ & 78.63$_{\textcolor{blue}{+1.68}}$ & 76.84$_{\textcolor{blue}{+1.28}}$ & 75.74$_{\textcolor{blue}{+1.32}}$ & 47.25$_{\textcolor{blue}{+6.08}}$ & 48.93$_{\textcolor{blue}{+3.47}}$ & 66.70$_{\textcolor{blue}{+1.78}}$ & 52.23$_{\textcolor{blue}{+5.54}}$ & 85.09$_{\textcolor{blue}{+0.61}}$ & 53.42$_{\textcolor{blue}{+0.81}}$ & 58.94$_{\textcolor{blue}{+3.05}}$
\\
T5 E-Ext (L) &  92.74$_{\textcolor{blue}{+0.35}}$ & 60.50$_{\textcolor{blue}{+0.88}}$ & 70.50$_{\textcolor{blue}{+0.28}}$ & 79.14$_{\textcolor{blue}{+0.62}}$ & 81.28$_{\textcolor{blue}{+1.22}}$ & 78.44$_{\textcolor{blue}{+0.51}}$ & 77.10$_{\textcolor{blue}{+0.64}}$ & 58.68$_{\textcolor{blue}{+5.95}}$ & 53.07$_{\textcolor{blue}{+1.69}}$ & 69.66$_{\textcolor{red}{-0.33}}$ & 55.04$_{\textcolor{blue}{+5.28}}$ & 86.53$_{\textcolor{blue}{+0.75}}$ & 55.28$_{\textcolor{blue}{+0.46}}$ & 63.04$_{\textcolor{blue}{+2.30}}$
\\
T5 ED-Ext (B) & 91.03$_{\textcolor{blue}{+0.46}}$ & 60.73$_{\textcolor{blue}{+2.73}}$ & 68.80$_{\textcolor{blue}{+1.93}}$ & 78.10$_{\textcolor{blue}{+0.44}}$ & 79.66$_{\textcolor{blue}{+0.98}}$ & 77.19$_{\textcolor{blue}{+0.50}}$ & 75.92$_{\textcolor{blue}{+1.17}}$ & 48.67$_{\textcolor{blue}{+3.18}}$ & 49.06$_{\textcolor{blue}{+3.50}}$ & 67.46$_{\textcolor{blue}{+0.47}}$ & 50.66$_{\textcolor{blue}{+2.00}}$ & 85.49$_{\textcolor{blue}{+0.58}}$ & 54.05$_{\textcolor{blue}{+3.02}}$ & 59.23$_{\textcolor{blue}{+2.12}}$
\\

T5 ED-Gen (B) & 91.29$_{\textcolor{blue}{+0.66}}$ & 66.37$_{\textcolor{red}{-0.37}}$ & 73.99$_{\textcolor{blue}{+0.54}}$ & 82.75$_{\textcolor{red}{0.00}}$ & 78.58$_{\textcolor{red}{-0.23}}$ & 75.41$_{\textcolor{blue}{+0.31}}$ & 78.06$_{\textcolor{blue}{+0.15}}$ & 51.13$_{\textcolor{blue}{+3.06}}$ & 48.99$_{\textcolor{blue}{+1.45}}$ & 68.65$_{\textcolor{blue}{+1.32}}$ & 47.09$_{\textcolor{blue}{+0.90}}$ & 85.84$_{\textcolor{blue}{+0.90}}$ & 45.23$_{\textcolor{blue}{+1.74}}$ & 57.82$_{\textcolor{blue}{+1.56}}$
\\ 
BART E-Ext (L) &
92.42$_{\textcolor{blue}{+0.27}}$ & 61.83$_{\textcolor{red}{-0.48}}$ & 70.98$_{\textcolor{red}{-1.86}}$ & 80.12$_{\textcolor{blue}{+0.13}}$ & 82.02$_{\textcolor{blue}{+1.50}}$ & 79.13$_{\textcolor{blue}{+0.27}}$ & 77.75$_{\textcolor{red}{-0.03}}$ & 58.32$_{\textcolor{blue}{+7.41}}$ & 50.06$_{\textcolor{blue}{+1.23}}$ & 69.62$_{\textcolor{blue}{+1.44}}$ & 55.02$_{\textcolor{blue}{+7.83}}$ & 86.79$_{\textcolor{blue}{+0.75}}$ & 59.83$_{\textcolor{blue}{+2.94}}$ & 63.27$_{\textcolor{blue}{+3.60}}$
\\
BART ED-Ext (L) & 
93.06$_{\textcolor{blue}{+0.56}}$ & 58.72$_{\textcolor{red}{-0.09}}$ & 70.80$_{\textcolor{red}{-1.31}}$ & 80.11$_{\textcolor{red}{-0.22}}$ & 81.78$_{\textcolor{blue}{+1.48}}$ & 79.11$_{\textcolor{blue}{+0.54}}$ & 77.26$_{\textcolor{blue}{+0.16}}$ & 60.19$_{\textcolor{blue}{+5.45}}$ & 48.97$_{\textcolor{blue}{+1.84}}$ & 67.47$_{\textcolor{blue}{+1.42}}$ & 53.24$_{\textcolor{blue}{+6.24}}$ & 86.75$_{\textcolor{blue}{+0.60}}$ & 56.22$_{\textcolor{blue}{+1.56}}$ & 62.14$_{\textcolor{blue}{+2.85}}$
\\
BART ED-Gen (L) & 
88.58$_{\textcolor{blue}{+9.86}}$ & 66.18$_{\textcolor{blue}{+3.00}}$ & 75.21$_{\textcolor{blue}{+5.99}}$ & 83.38$_{\textcolor{blue}{+3.99}}$ & 79.88$_{\textcolor{blue}{+7.16}}$ & 75.41$_{\textcolor{blue}{+19.32}}$ & 78.11$_{\textcolor{blue}{+8.22}}$ & 55.07$_{\textcolor{blue}{+11.03}}$ & 49.91$_{\textcolor{blue}{+6.27}}$ & 63.69$_{\textcolor{blue}{+9.90}}$ & 46.75$_{\textcolor{blue}{+8.31}}$ & 80.94$_{\textcolor{blue}{+8.77}}$ & 48.11$_{\textcolor{blue}{+2.27}}$ & 57.41$_{\textcolor{blue}{+7.76}}$

\\
ELECTRA (L) &
93.27$_{\textcolor{red}{-0.12}}$ & 60.59$_{\textcolor{blue}{+0.36}}$ & 72.96$_{\textcolor{red}{-3.35}}$ & 82.03$_{\textcolor{red}{-0.51}}$ & 83.10$_{\textcolor{blue}{+2.11}}$ & 79.16$_{\textcolor{blue}{+0.38}}$ & 78.52$_{\textcolor{red}{-0.19}}$ & 62.56$_{\textcolor{blue}{+7.13}}$ & 50.29$_{\textcolor{blue}{+0.49}}$ & 71.50$_{\textcolor{blue}{+4.54}}$ & 54.60$_{\textcolor{blue}{+6.80}}$ & 87.14$_{\textcolor{blue}{+0.91}}$ & 56.88$_{\textcolor{blue}{+1.98}}$ & 63.83$_{\textcolor{blue}{+3.64}}$
\\
RoBERTa (L) & 
93.36$_{\textcolor{blue}{+0.69}}$ & 60.15$_{\textcolor{blue}{+0.83}}$ & 71.40$_{\textcolor{red}{-1.12}}$ & 80.56$_{\textcolor{red}{-0.78}}$ & 82.21$_{\textcolor{blue}{+1.33}}$ & 79.50$_{\textcolor{blue}{+0.68}}$ & 77.86$_{\textcolor{blue}{+0.27}}$ & 64.79$_{\textcolor{blue}{+9.77}}$ & 51.49$_{\textcolor{blue}{+3.31}}$ & 68.69$_{\textcolor{blue}{+4.03}}$ & 53.68$_{\textcolor{blue}{+1.26}}$ & 86.31$_{\textcolor{red}{-0.34}}$ & 57.22$_{\textcolor{blue}{+2.24}}$ & 63.70$_{\textcolor{blue}{+3.38}}$
\\
\bottomrule 
\end{tabular}
}
}
\caption{Three readers trained by single and multi task learning and evaluated on in-domain and out-domain datasets by F1 Score. Inference length for all readers is 512. }
\label{tab:table_512}
\end{table*}
\begin{table*}[t]
\large
\centering
\renewcommand{\arraystretch}{1.2}
{
\resizebox{\linewidth}{!}{
\begin{tabular}{c|lllllll|lllllll}
\toprule

\multirow{2}{*}{Model} &  \multicolumn{7}{c}{In-domain Datasets} & \multicolumn{7}{|c}{Out-domain Datasets}\\
\cmidrule(lr){2-8} \cmidrule(lr){9-15}
~& SQuAD & NewsQA & TQA & SQA & HQA & NQ & Avg. & DROP & RACE & BioASQ & TbQA & RE & DuoRC & Avg. \\ 
 \toprule
    \multicolumn{3}{l}{\bf Single Task Learning}\\
    \bottomrule

T5 E-Ext (B) & 90.20 & 69.93 & 66.26 & 74.56 & 77.38 & 76.44 & 75.80 & 41.36 & 45.63 & 66.64 & 54.34 & 84.48 & 55.93 & 58.06\\

T5 E-Ext (L) & 92.47 & 72.22 & 70.43 & 77.10 & 80.69 & 79.08 & 78.67 & 53.14 & 52.06 & 71.26 & 61.07 & 85.78 & 58.72 & 63.67\\

T5 ED-Ext (B) & 90.71 & 70.43 & 68.48 & 76.01 & 78.94 & 77.80 & 77.06 & 45.86 & 46.18 & 67.93 & 55.07 & 84.91 & 55.19 & 59.19\\

T5 ED-Gen (B) & 90.75 & 71.64 & 79.02 & 86.09 & 79.87 & 76.72 & 80.68 & 48.08 & 48.89 & 67.36 & 60.42 & 84.94 & 60.83 & 61.75\\

BART E-Ext (L) & 92.19 & 72.20 & 73.12 & 77.19 & 80.61 & 79.29 & 79.10 & 51.57 & 48.82 & 68.83 & 51.29 & 86.04 & 61.35 & 61.32\\

BART ED-Ext (L) & 92.51 & 58.68 & 72.55 & 80.94 & 80.71 & 78.63 & 77.34 & 54.73 & 47.64 & 66.15 & 46.18 & 86.15 & 54.39 & 59.21\\

BART ED-Gen (L) & 78.75 & 66.20 & 67.81 & 78.89 & 73.22 & 56.58 & 70.24 & 44.22 & 43.70 & 55.59 & 45.11 & 76.83 & 55.63 & 53.51 \\

ELECTRA (L) & 93.39 & 60.23 & 76.31 & 82.54 & 80.99 & 78.78 & 78.71 & 55.43 & 49.80 & 66.96 & 47.80 & 86.23 & 54.90 & 60.19\\

RoBERTa (L) & 92.64 & 59.95 & 72.97 & 81.62 & 81.21 & 78.95 & 77.89 & 55.88 & 47.72 & 64.47 & 52.31 & 86.69 & 55.75 & 60.47\\

\toprule
\multicolumn{3}{l}{\bf Multi-Task Learning}\\
\bottomrule
T5 E-Ext (B) & 90.81$_{\textcolor{blue}{+0.61}}$ & 70.73$_{\textcolor{blue}{+0.80}}$ & 66.73$_{\textcolor{blue}{+0.47}}$ & 74.96$_{\textcolor{blue}{+0.40}}$ & 79.02$_{\textcolor{blue}{+1.64}}$ & 77.64$_{\textcolor{blue}{+1.20}}$ & 76.65$_{\textcolor{blue}{+0.85}}$ & 47.99$_{\textcolor{blue}{+6.63}}$ & 49.22$_{\textcolor{blue}{+3.59}}$ & 67.59$_{\textcolor{blue}{+0.95}}$ & 60.18$_{\textcolor{blue}{+5.84}}$ & 85.09$_{\textcolor{blue}{+0.61}}$ & 59.24$_{\textcolor{blue}{+3.31}}$ & 61.55$_{\textcolor{blue}{+3.49}}$
 \\
T5 E-Ext (L) &  92.84$_{\textcolor{blue}{+0.37}}$ & 73.15$_{\textcolor{blue}{+0.93}}$ & 70.86$_{\textcolor{blue}{+0.43}}$ & 77.30$_{\textcolor{blue}{+0.20}}$ & 81.88$_{\textcolor{blue}{+1.19}}$ & 79.77$_{\textcolor{blue}{+0.69}}$ & 79.30$_{\textcolor{blue}{+0.63}}$ & 59.10$_{\textcolor{blue}{+5.96}}$ & 54.01$_{\textcolor{blue}{+1.95}}$ & 71.13$_{\textcolor{red}{-0.13}}$ & 64.63$_{\textcolor{blue}{+3.56}}$ & 86.53$_{\textcolor{blue}{+0.75}}$ & 61.21$_{\textcolor{blue}{+2.49}}$ & 66.10$_{\textcolor{blue}{+2.43}}$
\\
T5 ED-Ext (B) & 91.12$_{\textcolor{blue}{+0.41}}$ & 71.78$_{\textcolor{blue}{+1.35}}$ & 66.93$_{\textcolor{red}{-1.55}}$ & 76.13$_{\textcolor{blue}{+0.12}}$ & 80.23$_{\textcolor{blue}{+1.29}}$ & 78.11$_{\textcolor{blue}{+0.31}}$ & 77.38$_{\textcolor{blue}{+0.32}}$ & 49.69$_{\textcolor{blue}{+3.83}}$ & 49.64$_{\textcolor{blue}{+3.46}}$ & 68.45$_{\textcolor{blue}{+0.52}}$ & 60.50$_{\textcolor{blue}{+5.43}}$ & 85.49$_{\textcolor{blue}{+0.58}}$ & 57.41$_{\textcolor{blue}{+2.22}}$ & 61.86$_{\textcolor{blue}{+2.67}}$
\\
T5 ED-Gen (B) & 91.41$_{\textcolor{blue}{+0.66}}$ & 71.27$_{\textcolor{red}{-0.37}}$ & 79.65$_{\textcolor{blue}{+0.63}}$ & 86.21$_{\textcolor{blue}{+0.12}}$ & 79.70$_{\textcolor{red}{-0.17}}$ & 77.10$_{\textcolor{blue}{+0.38}}$ & 80.89$_{\textcolor{blue}{+0.21}}$ & 51.20$_{\textcolor{blue}{+3.12}}$ & 49.66$_{\textcolor{blue}{+0.77}}$ & 68.72$_{\textcolor{blue}{+1.36}}$ & 63.02$_{\textcolor{blue}{+2.60}}$ & 85.84$_{\textcolor{blue}{+0.90}}$ & 62.94$_{\textcolor{blue}{+2.11}}$ & 63.56$_{\textcolor{blue}{+1.81}}$\\ 
BART E-Ext (L) & 
92.46$_{\textcolor{blue}{+0.27}}$ & 72.11$_{\textcolor{red}{-0.09}}$ & 72.24$_{\textcolor{red}{-0.88}}$ & 76.53$_{\textcolor{red}{-0.66}}$ & 82.04$_{\textcolor{blue}{+1.43}}$ & 79.40$_{\textcolor{blue}{+0.11}}$ & 79.13$_{\textcolor{blue}{+0.03}}$ & 58.22$_{\textcolor{blue}{+6.65}}$ & 50.40$_{\textcolor{blue}{+1.58}}$ & 70.72$_{\textcolor{blue}{+1.89}}$ & 56.29$_{\textcolor{blue}{+5.00}}$ & 86.79$_{\textcolor{blue}{+0.75}}$ & 61.95$_{\textcolor{blue}{+0.60}}$ & 64.06$_{\textcolor{blue}{+2.74}}$
\\

BART ED-Ext (L) & 93.07$_{\textcolor{blue}{+0.56}}$ & 58.67$_{\textcolor{red}{-0.01}}$ & 71.47$_{\textcolor{red}{-1.08}}$ & 80.66$_{\textcolor{red}{-0.28}}$ & 82.14$_{\textcolor{blue}{+1.43}}$ & 79.32$_{\textcolor{blue}{+0.69}}$ & 77.55$_{\textcolor{blue}{+0.21}}$ & 60.40$_{\textcolor{blue}{+5.67}}$ & 51.32$_{\textcolor{blue}{+3.68}}$ & 67.48$_{\textcolor{blue}{+1.33}}$ & 53.34$_{\textcolor{blue}{+7.16}}$ & 86.75$_{\textcolor{blue}{+0.60}}$ & 56.79$_{\textcolor{blue}{+2.40}}$ & 62.68$_{\textcolor{blue}{+3.47}}$
\\
BART ED-Gen (L) & -
88.63$_{\textcolor{blue}{+9.88}}$ & 68.91$_{\textcolor{blue}{+2.71}}$ & 74.91$_{\textcolor{blue}{+7.10}}$ & 82.52$_{\textcolor{blue}{+3.63}}$ & 80.53$_{\textcolor{blue}{+7.31}}$ & 75.78$_{\textcolor{blue}{+19.20}}$ & 78.55$_{\textcolor{blue}{+8.31}}$ & 55.20$_{\textcolor{blue}{+10.98}}$ & 50.04$_{\textcolor{blue}{+6.34}}$ & 63.78$_{\textcolor{blue}{+8.19}}$ & 54.81$_{\textcolor{blue}{+9.70}}$ & 80.94$_{\textcolor{blue}{+4.11}}$ & 58.47$_{\textcolor{blue}{+2.84}}$ & 60.54$_{\textcolor{blue}{+7.03}}$
\\
ELECTRA (L) & 93.27$_{\textcolor{red}{-0.12}}$ & 60.59$_{\textcolor{blue}{+0.36}}$ & 72.96$_{\textcolor{red}{-3.35}}$ & 82.03$_{\textcolor{red}{-0.51}}$ & 83.10$_{\textcolor{blue}{+2.11}}$ & 79.16$_{\textcolor{blue}{+0.38}}$ & 78.52$_{\textcolor{red}{-0.19}}$ & 62.56$_{\textcolor{blue}{+7.13}}$ & 50.29$_{\textcolor{blue}{+0.49}}$ & 71.50$_{\textcolor{blue}{+4.54}}$ & 54.60$_{\textcolor{blue}{+6.80}}$ & 87.14$_{\textcolor{blue}{+0.91}}$ & 56.88$_{\textcolor{blue}{+1.98}}$ & 63.83$_{\textcolor{blue}{+3.64}}$\\

RoBERTa (L) & 
93.41$_{\textcolor{blue}{+0.77}}$ & 59.56$_{\textcolor{red}{-0.39}}$ & 72.23$_{\textcolor{red}{-0.74}}$ & 80.98$_{\textcolor{red}{-0.64}}$ & 82.37$_{\textcolor{blue}{+1.16}}$ & 79.55$_{\textcolor{blue}{+0.60}}$ & 78.02$_{\textcolor{blue}{+0.13}}$ & 64.47$_{\textcolor{blue}{+8.59}}$ & 51.81$_{\textcolor{blue}{+4.09}}$ & 69.15$_{\textcolor{blue}{+4.68}}$ & 53.68$_{\textcolor{blue}{+1.37}}$ & 86.31$_{\textcolor{red}{-0.38}}$ & 56.06$_{\textcolor{blue}{+0.31}}$ & 63.58$_{\textcolor{blue}{+3.11}}$\\

\bottomrule 
\end{tabular}
}
}
\caption{Three readers trained by single and multi task learning and evaluated on in-domain and out-domain datasets by F1 Score. Inference length for all readers is 1024, except for ELECTRA is 512.}
\label{tab:table_1024}
\end{table*}
\begin{table*}[t]
\large
\centering
\renewcommand{\arraystretch}{1.2}
{
\resizebox{\linewidth}{!}{
\begin{tabular}{c|lllllll|lllllll}
\toprule

\multirow{2}{*}{Model} &  \multicolumn{7}{c}{In-domain Datasets} & \multicolumn{7}{|c}{Out-of-domain Datasets}\\
\cmidrule(lr){2-8} \cmidrule(lr){9-15}
~& SQuAD & NewsQA & TQA & SQA & HQA & NQ & Avg. & DROP & RACE & BioASQ & TbQA & RE & DuoRC & Avg. \\ 
\toprule
\multicolumn{3}{l}{\bf Single Task Learning}\\
\bottomrule
T5 E-Ext (B) & 90.20 & 70.14 & 72.67 & 79.89 & 77.37 & 77.31 & 77.93 & 41.36 & 45.63 & 66.64 & 55.17 & 84.48 & 58.62 & 58.65\\
T5 E-Ext (L) &92.47 & 72.63 & 76.09 & 83.24 & 80.67 & 80.00 & 80.85 & 53.14 & 52.06 & 71.26 & 61.92 & 85.78 & 62.80 & 64.49\\
T5 ED-Ext (B) & 90.71 & 70.80 & 74.16 & 81.32 & 78.98 & 78.68 & 79.11 & 45.86 & 46.18 & 67.93 & 55.74 & 84.91 & 59.33 & 59.99\\
T5 ED-Gen (B) & 90.75 & 71.65 & {79.61} & {86.21} & 79.89 & 78.04 & {81.02} & 48.08 & 48.89 & 67.36 & 60.30 & 84.94 & 61.35 & 61.82\\
BART E-Ext (L) & 
92.19 & 72.20 & 73.12 & 77.19 & 80.61 & 79.29 & 79.10 & 51.57 & 48.82 & 68.83 & 51.29 & 86.04 & 61.35 & 61.32\\
BART ED-Ext (L)  & 92.51 & 58.68 & 72.55 & 80.94 & 80.71 & 78.63 & 77.34 & 54.73 & 47.64 & 66.15 & 46.18 & 86.15 & 54.39 & 59.21\\
BART ED-Gen (L) & 
78.75 & 66.20 & 67.81 & 78.89 & 73.22 & 56.58 & 70.24 & 44.22 & 43.70 & 55.59 & 45.11 & 76.83 & 55.63 & 53.51 \\
ELECTRA (L) & 93.39 & 60.23 & 76.31 & 82.54 & 80.99 & 78.78 & 78.71 & 55.43 & 49.80 & 66.96 & 47.80 & 86.23 & 54.90 & 60.19\\
RoBERTa (L) & 92.64 & 59.95 & 72.97 & 81.62 & 81.21 & 78.95 & 77.89 & 55.88 & 47.72 & 64.47 & 52.31 & 86.69 & 55.75 & 60.47\\
\toprule
\multicolumn{3}{l}{\bf Multi-Task Learning}\\
\bottomrule
T5 E-Ext (B) & 
90.81$_{\textcolor{blue}{+0.61}}$ & 70.92$_{\textcolor{blue}{+0.78}}$ & 74.22$_{\textcolor{blue}{+1.55}}$ & 80.42$_{\textcolor{blue}{+0.53}}$ & 79.03$_{\textcolor{blue}{+1.66}}$ & 78.57$_{\textcolor{blue}{+1.26}}$ & 78.99$_{\textcolor{blue}{+1.06}}$ & 47.99$_{\textcolor{blue}{+6.63}}$ & 49.22$_{\textcolor{blue}{+3.59}}$ & 67.59$_{\textcolor{blue}{+0.95}}$ & 60.52$_{\textcolor{blue}{+5.35}}$ & 85.09$_{\textcolor{blue}{+0.61}}$ & 61.44$_{\textcolor{blue}{+2.82}}$ & 61.98$_{\textcolor{blue}{+3.33}}$ \\
T5 E-Ext (L) & 92.84$_{\textcolor{blue}{+0.37}}$ & 73.51$_{\textcolor{blue}{+0.88}}$ & 77.37$_{\textcolor{blue}{+1.28}}$ & 82.89$_{\textcolor{red}{-0.35}}$ & 81.92$_{\textcolor{blue}{+1.25}}$ & 80.74$_{\textcolor{blue}{+0.74}}$ & 81.55$_{\textcolor{blue}{+0.70}}$ & 59.10$_{\textcolor{blue}{+5.96}}$ & 54.01$_{\textcolor{blue}{+1.95}}$ & 71.13$_{\textcolor{red}{-0.13}}$ & 64.90$_{\textcolor{blue}{+2.98}}$ & 86.53$_{\textcolor{blue}{+0.75}}$ & 65.01$_{\textcolor{blue}{+2.21}}$ & 66.78$_{\textcolor{blue}{+2.29}}$
\\
T5 ED-Ext (B) & 91.12$_{\textcolor{blue}{+0.41}}$ & 71.95$_{\textcolor{blue}{+1.15}}$ & 75.50$_{\textcolor{blue}{+1.34}}$ & 81.82$_{\textcolor{blue}{+0.50}}$ & 80.25$_{\textcolor{blue}{+1.27}}$ & 78.93$_{\textcolor{blue}{+0.25}}$ & 79.93$_{\textcolor{blue}{+0.82}}$ & 49.69$_{\textcolor{blue}{+3.83}}$ & 49.64$_{\textcolor{blue}{+3.46}}$ & 68.45$_{\textcolor{blue}{+0.52}}$ & 61.33$_{\textcolor{blue}{+5.59}}$ & 85.49$_{\textcolor{blue}{+0.58}}$ & 61.22$_{\textcolor{blue}{+1.89}}$ & 62.64$_{\textcolor{blue}{+2.65}}$
\\
T5 ED-Gen (L)  & 91.41$_{\textcolor{blue}{+0.66}}$ & 71.29$_{\textcolor{red}{-0.36}}$ & {80.01}$_{\textcolor{blue}{+0.40}}$ & {86.46}$_{\textcolor{blue}{+0.25}}$ & 79.70$_{\textcolor{red}{-0.19}}$ & 78.09$_{\textcolor{blue}{+0.05}}$ & 81.16$_{\textcolor{blue}{+0.14}}$ & 51.20$_{\textcolor{blue}{+3.12}}$ & 49.66$_{\textcolor{blue}{+0.77}}$ & 68.72$_{\textcolor{blue}{+1.36}}$ & 62.90$_{\textcolor{blue}{+2.60}}$ & 85.84$_{\textcolor{blue}{+0.90}}$ & 63.76$_{\textcolor{blue}{+2.41}}$ & 63.68$_{\textcolor{blue}{+1.86}}$\\ 

BART E-Ext (L) & 
92.46$_{\textcolor{blue}{+0.27}}$ & 72.11$_{\textcolor{red}{-0.09}}$ & 72.24$_{\textcolor{red}{-0.88}}$ & 76.53$_{\textcolor{red}{-0.66}}$ & 82.04$_{\textcolor{blue}{+1.43}}$ & 79.40$_{\textcolor{blue}{+0.11}}$ & 79.13$_{\textcolor{blue}{+0.03}}$ & 58.22$_{\textcolor{blue}{+6.65}}$ & 50.40$_{\textcolor{blue}{+1.58}}$ & 70.72$_{\textcolor{blue}{+1.89}}$ & 56.29$_{\textcolor{blue}{+5.00}}$ & 86.79$_{\textcolor{blue}{+0.75}}$ & 61.95$_{\textcolor{blue}{+0.60}}$ & 64.06$_{\textcolor{blue}{+2.74}}$
\\
BART ED-Ext (L) & 
93.07$_{\textcolor{blue}{+0.56}}$ & 58.67$_{\textcolor{red}{-0.01}}$ & 71.47$_{\textcolor{red}{-1.08}}$ & 80.66$_{\textcolor{red}{-0.28}}$ & 82.14$_{\textcolor{blue}{+1.43}}$ & 79.32$_{\textcolor{blue}{+0.69}}$ & 77.55$_{\textcolor{blue}{+0.21}}$ & 60.40$_{\textcolor{blue}{+5.67}}$ & 51.32$_{\textcolor{blue}{+3.68}}$ & 67.48$_{\textcolor{blue}{+1.33}}$ & 53.34$_{\textcolor{blue}{+7.16}}$ & 86.75$_{\textcolor{blue}{+0.60}}$ & 56.79$_{\textcolor{blue}{+2.40}}$ & 62.68$_{\textcolor{blue}{+3.47}}$\\

BART ED-Gen (L)& 
88.63$_{\textcolor{blue}{+9.88}}$ & 68.91$_{\textcolor{blue}{+2.71}}$ & 74.91$_{\textcolor{blue}{+7.10}}$ & 82.52$_{\textcolor{blue}{+3.63}}$ & 80.53$_{\textcolor{blue}{+7.31}}$ & 75.78$_{\textcolor{blue}{+19.20}}$ & 78.55$_{\textcolor{blue}{+8.31}}$ & 55.20$_{\textcolor{blue}{+10.98}}$ & 50.04$_{\textcolor{blue}{+6.34}}$ & 63.78$_{\textcolor{blue}{+8.19}}$ & 54.81$_{\textcolor{blue}{+9.70}}$ & 80.94$_{\textcolor{blue}{+4.11}}$ & 58.47$_{\textcolor{blue}{+2.84}}$ & 60.54$_{\textcolor{blue}{+7.03}}$\\

ELECTRA (L)& 93.27$_{\textcolor{red}{-0.12}}$ & 60.59$_{\textcolor{blue}{+0.36}}$ & 72.96$_{\textcolor{red}{-3.35}}$ & 82.03$_{\textcolor{red}{-0.51}}$ & {83.10}$_{\textcolor{blue}{+2.11}}$ & 79.16$_{\textcolor{blue}{+0.38}}$ & 78.52$_{\textcolor{red}{-0.19}}$ & 62.56$_{\textcolor{blue}{+7.13}}$ & 50.29$_{\textcolor{blue}{+0.49}}$ & {71.50}$_{\textcolor{blue}{+4.54}}$ & 54.60$_{\textcolor{blue}{+6.80}}$ & {87.14}$_{\textcolor{blue}{+0.91}}$ & 56.88$_{\textcolor{blue}{+1.98}}$ & 63.83$_{\textcolor{blue}{+3.64}}$\\
RoBERTa (L)& 
{93.41}$_{\textcolor{blue}{+0.77}}$ & 59.56$_{\textcolor{red}{-0.39}}$ & 72.23$_{\textcolor{red}{-0.74}}$ & 80.98$_{\textcolor{red}{-0.64}}$ & 82.37$_{\textcolor{blue}{+1.16}}$ & 79.55$_{\textcolor{blue}{+0.60}}$ & 78.02$_{\textcolor{blue}{+0.13}}$ & {64.47}$_{\textcolor{blue}{+8.59}}$ & 51.81$_{\textcolor{blue}{+4.09}}$ & 69.15$_{\textcolor{blue}{+4.68}}$ & 53.68$_{\textcolor{blue}{+1.37}}$ & 86.31$_{\textcolor{red}{-0.38}}$ & 56.06$_{\textcolor{blue}{+0.31}}$ & 63.58$_{\textcolor{blue}{+3.11}}$\\
\bottomrule 
\end{tabular}
}
}
\caption{Three readers trained by single and multi task learning and evaluated on in-domain and out-domain datasets by F1 Score.  Inference length for T5 readers is full length, for BART is 1024, and for ELECTRA is 512.}
\label{tab:table_max}
\end{table*}

\section{Detailed Comparison Results for Using Same PrLMs} \label{apd:same_prlms}
Table \ref{tab:same_prlms} presents the F1 score of each readers when using the same PrLMs as we discussed in \S\ref{sec:comp_same_prlms}.
\begin{table*}[t]
\large
\centering
\renewcommand{\arraystretch}{1.2}
{
\resizebox{\linewidth}{!}{
\begin{tabular}{c|lllllll|lllllll}
\toprule

\multirow{2}{*}{Model} &  \multicolumn{7}{c}{In-domain Datasets} & \multicolumn{7}{|c}{Out-of-domain Datasets}\\
\cmidrule(lr){2-8} \cmidrule(lr){9-15}
~& SQuAD & NewsQA & TQA & SQA & HQA & NQ & Avg. & DROP & RACE & BioASQ & TbQA & RE & DuoRC & Avg. \\ 
\toprule
\multicolumn{3}{l}{\bf Single Task Learning}\\
\bottomrule
T5 E-Ext (B) & 90.20 & 70.14 & 72.67 & 79.89 & 77.37 & 77.31 & 77.93 & 41.36 & 45.63 & 66.64 & 55.17 & 84.48 & 58.62 & 58.65\\
T5 ED-Ext (B) & 90.71 & 70.80 & 74.16 & 81.32 & 78.98 & 78.68 & 79.11 & 45.86 & 46.18 & 67.93 & 55.74 & 84.91 & 59.33 & 59.99\\
T5 ED-Gen (B) & 90.75 & 71.65 & {79.61} & {86.21} & 79.89 & 78.04 & {81.02} & 48.08 & 48.89 & 67.36 & 60.30 & 84.94 & 61.35 & 61.82\\
BART E-Ext (L) & 
92.19 & 72.20 & 73.12 & 77.19 & 80.61 & 79.29 & 79.10 & 51.57 & 48.82 & 68.83 & 51.29 & 86.04 & 61.35 & 61.32\\
BART ED-Ext  & 92.51 & 58.68 & 72.55 & 80.94 & 80.71 & 78.63 & 77.34 & 54.73 & 47.64 & 66.15 & 46.18 & 86.15 & 54.39 & 59.21\\
BART ED-Gen (L) & 
78.75 & 66.20 & 67.81 & 78.89 & 73.22 & 56.58 & 70.24 & 44.22 & 43.70 & 55.59 & 45.11 & 76.83 & 55.63 & 53.51 \\
\toprule
\multicolumn{3}{l}{\bf Multi-Task Learning}\\
\bottomrule
T5 E-Ext (B) & 
90.81$_{\textcolor{blue}{+0.61}}$ & 70.92$_{\textcolor{blue}{+0.78}}$ & 74.22$_{\textcolor{blue}{+1.55}}$ & 80.42$_{\textcolor{blue}{+0.53}}$ & 79.03$_{\textcolor{blue}{+1.66}}$ & 78.57$_{\textcolor{blue}{+1.26}}$ & 78.99$_{\textcolor{blue}{+1.06}}$ & 47.99$_{\textcolor{blue}{+6.63}}$ & 49.22$_{\textcolor{blue}{+3.59}}$ & 67.59$_{\textcolor{blue}{+0.95}}$ & 60.52$_{\textcolor{blue}{+5.35}}$ & 85.09$_{\textcolor{blue}{+0.61}}$ & 61.44$_{\textcolor{blue}{+2.82}}$ & 61.98$_{\textcolor{blue}{+3.33}}$ \\
T5 ED-Ext (B) & 91.12$_{\textcolor{blue}{+0.41}}$ & 71.95$_{\textcolor{blue}{+1.15}}$ & 75.50$_{\textcolor{blue}{+1.34}}$ & 81.82$_{\textcolor{blue}{+0.50}}$ & 80.25$_{\textcolor{blue}{+1.27}}$ & 78.93$_{\textcolor{blue}{+0.25}}$ & 79.93$_{\textcolor{blue}{+0.82}}$ & 49.69$_{\textcolor{blue}{+3.83}}$ & 49.64$_{\textcolor{blue}{+3.46}}$ & 68.45$_{\textcolor{blue}{+0.52}}$ & 61.33$_{\textcolor{blue}{+5.59}}$ & 85.49$_{\textcolor{blue}{+0.58}}$ & 61.22$_{\textcolor{blue}{+1.89}}$ & 62.64$_{\textcolor{blue}{+2.65}}$
\\
T5 ED-Gen (L)  & 91.41$_{\textcolor{blue}{+0.66}}$ & 71.29$_{\textcolor{red}{-0.36}}$ & {80.01}$_{\textcolor{blue}{+0.40}}$ & {86.46}$_{\textcolor{blue}{+0.25}}$ & 79.70$_{\textcolor{red}{-0.19}}$ & 78.09$_{\textcolor{blue}{+0.05}}$ & 81.16$_{\textcolor{blue}{+0.14}}$ & 51.20$_{\textcolor{blue}{+3.12}}$ & 49.66$_{\textcolor{blue}{+0.77}}$ & 68.72$_{\textcolor{blue}{+1.36}}$ & 62.90$_{\textcolor{blue}{+2.60}}$ & 85.84$_{\textcolor{blue}{+0.90}}$ & 63.76$_{\textcolor{blue}{+2.41}}$ & 63.68$_{\textcolor{blue}{+1.86}}$\\ 

BART E-Ext (L) & 
92.46$_{\textcolor{blue}{+0.27}}$ & 72.11$_{\textcolor{red}{-0.09}}$ & 72.24$_{\textcolor{red}{-0.88}}$ & 76.53$_{\textcolor{red}{-0.66}}$ & 82.04$_{\textcolor{blue}{+1.43}}$ & 79.40$_{\textcolor{blue}{+0.11}}$ & 79.13$_{\textcolor{blue}{+0.03}}$ & 58.22$_{\textcolor{blue}{+6.65}}$ & 50.40$_{\textcolor{blue}{+1.58}}$ & 70.72$_{\textcolor{blue}{+1.89}}$ & 56.29$_{\textcolor{blue}{+5.00}}$ & 86.79$_{\textcolor{blue}{+0.75}}$ & 61.95$_{\textcolor{blue}{+0.60}}$ & 64.06$_{\textcolor{blue}{+2.74}}$
\\
BART ED-Ext (L) & 
93.07$_{\textcolor{blue}{+0.56}}$ & 58.67$_{\textcolor{red}{-0.01}}$ & 71.47$_{\textcolor{red}{-1.08}}$ & 80.66$_{\textcolor{red}{-0.28}}$ & 82.14$_{\textcolor{blue}{+1.43}}$ & 79.32$_{\textcolor{blue}{+0.69}}$ & 77.55$_{\textcolor{blue}{+0.21}}$ & 60.40$_{\textcolor{blue}{+5.67}}$ & 51.32$_{\textcolor{blue}{+3.68}}$ & 67.48$_{\textcolor{blue}{+1.33}}$ & 53.34$_{\textcolor{blue}{+7.16}}$ & 86.75$_{\textcolor{blue}{+0.60}}$ & 56.79$_{\textcolor{blue}{+2.40}}$ & 62.68$_{\textcolor{blue}{+3.47}}$\\

BART ED-Gen (L)& 
88.63$_{\textcolor{blue}{+9.88}}$ & 68.91$_{\textcolor{blue}{+2.71}}$ & 74.91$_{\textcolor{blue}{+7.10}}$ & 82.52$_{\textcolor{blue}{+3.63}}$ & 80.53$_{\textcolor{blue}{+7.31}}$ & 75.78$_{\textcolor{blue}{+19.20}}$ & 78.55$_{\textcolor{blue}{+8.31}}$ & 55.20$_{\textcolor{blue}{+10.98}}$ & 50.04$_{\textcolor{blue}{+6.34}}$ & 63.78$_{\textcolor{blue}{+8.19}}$ & 54.81$_{\textcolor{blue}{+9.70}}$ & 80.94$_{\textcolor{blue}{+4.11}}$ & 58.47$_{\textcolor{blue}{+2.84}}$ & 60.54$_{\textcolor{blue}{+7.03}}$\\
\bottomrule 
\end{tabular}
}
}
\caption{Comparison of readers based on the same PrLMs by F1 Score. For three T5 readers, we use the T5-base model, for three BART readers, we use the BART-large model. Avg. means the Macro Average of in/out-domain datasets. Inference length for T5 is full length of context, for ELECTRA is 512 and for BART and RoBERTa is 1024.}
\label{tab:same_prlms}
\end{table*}

\begin{table*}[t]
\centering
\small
\renewcommand{\arraystretch}{1.2}
{
\resizebox{\linewidth}{!}{
\begin{tabular}{c|l|cccccc|cccccc}
\toprule

\multirow{2}{*}{Model} & \multirow{2}{*}{\diagbox{Train}{Test}} & \multicolumn{6}{c}{In-domain Datasets} & \multicolumn{6}{|c}{Out-domain Datasets}\\
\cmidrule(lr){3-8} \cmidrule(lr){9-14}
~& ~& SQuAD & NewsQA & TQA & SQA & HQA & NQ & DROP & RACE & BioASQ & TbQA & RE & DuoRC \\ 
 \hline
     \toprule
    \multicolumn{3}{l}{\bf Single Task Learning}\\
    \bottomrule
\multirow{6}{*}{T5 E-Ext (B)} 
& SQuAD &
90.20 & 63.37 & 63.75 & 30.97 & 67.53 & 62.28 & 36.03 & 45.63 & 66.38 & 54.77 & 84.48 & 57.08 \\
& NewsQA &
84.54 & 70.14 & 63.99 & 42.32 & 61.55 & 63.50 & 23.48 & 44.07 & 62.13 & 50.25 & 77.59 & 58.62 \\
& TQA &
69.68 & 46.83 & 72.67 & 60.40 & 54.33 & 54.49 & 24.28 & 37.15 & 60.07 & 42.61 & 75.83 & 47.72 \\
& SQA &
60.75 & 40.49 & 68.37 & 79.89 & 44.21 & 49.84 & 23.68 & 30.02 & 55.93 & 39.28 & 75.26 & 43.36 \\
& HQA &
83.30 & 59.19 & 61.67 & 48.18 & 77.37 & 62.35 & 39.04 & 40.51 & 63.68 & 40.15 & 84.07 & 55.31 \\
& NQ &
83.87 & 60.81 & 65.64 & 52.24 & 64.60 & 77.31 & 41.36 & 43.99 & 66.64 & 55.17 & 82.58 & 52.88 \\
 \hline
 \multirow{6}{*}{T5 E-Ext (L)} 
& SQuAD &
92.47 & 65.33 & 67.97 & 32.73 & 71.00 & 64.97 & 52.01 & 50.13 & 68.66 & 53.03 & 85.78 & 61.41 \\
& NewsQA &
87.38 & 72.63 & 69.34 & 43.83 & 66.56 & 69.02 & 31.72 & 49.72 & 65.97 & 55.51 & 78.75 & 62.80 \\
& TQA &
74.97 & 50.27 & 76.09 & 63.26 & 57.26 & 58.68 & 40.09 & 38.55 & 65.95 & 52.34 & 81.01 & 55.21 \\
& SQA &
72.47 & 48.12 & 73.57 & 83.24 & 53.50 & 57.17 & 41.57 & 35.53 & 66.07 & 52.64 & 81.63 & 52.05 \\
& HQA &
86.88 & 62.42 & 66.16 & 46.47 & 80.67 & 67.13 & 47.43 & 45.10 & 68.27 & 51.37 & 84.89 & 56.80 \\
& NQ &
86.73 & 64.62 & 70.32 & 54.09 & 68.54 & 80.00 & 53.14 & 52.06 & 71.26 & 61.92 & 84.35 & 60.43 \\
 \hline
 \multirow{6}{*}{T5 ED-Ext (B)} 
 & SQuAD &
92.47 & 65.33 & 67.97 & 32.73 & 71.00 & 64.97 & 52.01 & 50.13 & 68.66 & 53.03 & 85.78 & 61.41 \\
& NewsQA &
87.38 & 72.63 & 69.34 & 43.83 & 66.56 & 69.02 & 31.72 & 49.72 & 65.97 & 55.51 & 78.75 & 62.80 \\
& TQA &
74.97 & 50.27 & 76.09 & 63.26 & 57.26 & 58.68 & 40.09 & 38.55 & 65.95 & 52.34 & 81.01 & 55.21 \\
& SQA &
72.47 & 48.12 & 73.57 & 83.24 & 53.50 & 57.17 & 41.57 & 35.53 & 66.07 & 52.64 & 81.63 & 52.05 \\
& HQA &
86.88 & 62.42 & 66.16 & 46.47 & 80.67 & 67.13 & 47.43 & 45.10 & 68.27 & 51.37 & 84.89 & 56.80 \\
& NQ &
86.73 & 64.62 & 70.32 & 54.09 & 68.54 & 80.00 & 53.14 & 52.06 & 71.26 & 61.92 & 84.35 & 60.43 \\
 \hline
 \multirow{6}{*}{T5 ED-Gen (B)} 
 & SQuAD &
90.75 & 60.51 & 69.56 & 24.11 & 68.57 & 57.19 & 43.31 & 48.89 & 65.96 & 46.75 & 84.94 & 60.31 \\
& NewsQA &
85.75 & 71.65 & 69.70 & 43.16 & 63.61 & 62.96 & 25.37 & 45.97 & 62.80 & 53.82 & 77.37 & 61.35 \\
& TQA &
74.33 & 49.26 & 79.61 & 57.14 & 58.75 & 55.18 & 33.84 & 42.38 & 56.94 & 51.16 & 80.52 & 52.69 \\
& SQA &
70.62 & 44.66 & 78.03 & 86.21 & 57.19 & 52.92 & 35.32 & 35.33 & 59.76 & 53.66 & 79.54 & 49.23 \\
& HQA &
86.24 & 60.25 & 70.57 & 51.23 & 79.89 & 62.33 & 44.94 & 46.38 & 66.93 & 42.65 & 84.56 & 59.60 \\
& NQ &
85.46 & 61.80 & 72.08 & 57.55 & 67.71 & 78.04 & 48.08 & 45.85 & 67.36 & 60.30 & 84.06 & 58.42 \\
 \hline
 \multirow{6}{*}{BART E-Ext (L)} 
 & SQuAD &
92.19 & 62.30 & 60.86 & 35.52 & 69.60 & 62.94 & 51.31 & 48.82 & 68.83 & 49.39 & 86.04 & 58.31 \\
& NewsQA &
85.04 & 72.20 & 62.86 & 41.17 & 61.81 & 65.84 & 31.99 & 48.82 & 61.98 & 49.29 & 77.30 & 61.35 \\
& TQA &
68.36 & 43.38 & 73.12 & 55.53 & 59.27 & 55.11 & 37.79 & 36.16 & 53.90 & 37.98 & 80.07 & 49.51 \\
& SQA &
50.74 & 31.48 & 66.74 & 77.19 & 40.65 & 43.53 & 22.15 & 23.90 & 53.76 & 36.38 & 66.48 & 37.12 \\
& HQA &
82.21 & 52.46 & 56.53 & 34.95 & 80.61 & 62.58 & 44.30 & 39.60 & 59.40 & 33.74 & 85.46 & 52.60 \\
& NQ &
83.12 & 59.44 & 62.12 & 49.19 & 62.73 & 79.29 & 51.57 & 43.23 & 64.77 & 51.29 & 83.13 & 54.63 \\
 \hline
 \multirow{6}{*}{BART ED-Ext (L)} 
& SQuAD &
92.51 & 53.70 & 62.64 & 41.85 & 67.69 & 60.82 & 54.73 & 47.64 & 66.15 & 46.18 & 86.15 & 54.39 \\
& NewsQA &
86.15 & 58.68 & 62.29 & 46.98 & 64.09 & 66.00 & 31.91 & 45.52 & 60.70 & 44.82 & 78.72 & 54.09 \\
& TQA &
69.82 & 38.40 & 72.55 & 61.02 & 61.05 & 54.10 & 34.63 & 36.36 & 54.34 & 39.35 & 81.28 & 46.43 \\
& SQA &
57.26 & 32.09 & 69.35 & 80.94 & 41.82 & 45.62 & 28.54 & 25.18 & 51.50 & 41.09 & 70.98 & 38.88 \\
& HQA &
83.29 & 49.66 & 63.18 & 40.46 & 80.71 & 63.52 & 47.91 & 38.56 & 59.78 & 34.60 & 84.32 & 52.04 \\
& NQ &
83.86 & 50.35 & 64.06 & 56.34 & 62.53 & 78.63 & 52.41 & 44.25 & 65.59 & 45.93 & 84.43 & 49.44 \\
 \hline
 \multirow{6}{*}{BART ED-Gen (L)} 
& SQuAD &
78.75 & 54.02 & 48.69 & 22.33 & 57.19 & 57.90 & 44.09 & 41.33 & 47.04 & 35.42 & 70.68 & 45.79 \\
& NewsQA &
78.65 & 66.20 & 58.02 & 36.31 & 57.91 & 61.10 & 28.36 & 43.70 & 53.71 & 45.11 & 72.17 & 55.63 \\
& TQA &
58.98 & 39.22 & 67.81 & 53.90 & 54.81 & 46.73 & 32.85 & 33.74 & 46.62 & 39.97 & 64.89 & 45.47 \\
& SQA &
40.51 & 28.33 & 65.42 & 78.89 & 37.05 & 36.12 & 23.45 & 22.42 & 46.71 & 39.43 & 52.23 & 38.24 \\
& HQA &
74.75 & 50.41 & 56.56 & 40.90 & 73.22 & 57.83 & 44.22 & 37.31 & 55.59 & 29.96 & 76.83 & 50.62 \\
& NQ &
61.09 & 39.05 & 38.21 & 33.48 & 43.59 & 56.58 & 40.27 & 32.01 & 51.24 & 36.63 & 59.46 & 33.69 \\
 \hline
 \multirow{6}{*}{RoBERTa (L)} 
& SQuAD &
92.64 & 54.76 & 65.90 & 45.76 & 71.35 & 59.43 & 52.51 & 47.13 & 64.47 & 52.31 & 86.69 & 55.75 \\
& NewsQA &
86.50 & 59.95 & 63.01 & 48.02 & 66.99 & 67.29 & 33.52 & 47.26 & 60.05 & 45.10 & 78.08 & 54.27 \\
& TQA &
73.63 & 41.05 & 72.97 & 51.16 & 62.44 & 55.76 & 44.40 & 39.27 & 54.92 & 42.72 & 82.32 & 49.89 \\
& SQA &
53.59 & 29.57 & 70.35 & 81.62 & 42.03 & 47.06 & 23.04 & 23.70 & 54.18 & 39.69 & 71.13 & 36.06 \\
& HQA &
85.10 & 50.55 & 65.06 & 44.31 & 81.21 & 63.88 & 51.74 & 36.86 & 62.44 & 37.49 & 85.07 & 54.02 \\
& NQ &
85.25 & 49.49 & 64.48 & 57.23 & 67.47 & 78.95 & 55.88 & 47.72 & 63.77 & 44.67 & 84.10 & 50.00 \\
 \hline
 \multirow{6}{*}{ELECTRA (L)} 
& SQuAD &
93.39 & 55.42 & 65.92 & 46.56 & 68.69 & 68.92 & 55.11 & 49.80 & 66.96 & 46.57 & 86.23 & 54.90 \\
& NewsQA &
86.33 & 60.23 & 65.13 & 49.39 & 63.97 & 68.03 & 30.74 & 46.45 & 64.86 & 46.79 & 78.21 & 53.78 \\
& TQA &
69.75 & 40.20 & 76.31 & 65.27 & 58.87 & 55.95 & 42.21 & 37.46 & 59.94 & 41.54 & 80.56 & 49.24 \\
& SQA &
52.17 & 28.21 & 71.39 & 82.54 & 44.81 & 43.28 & 36.68 & 22.47 & 58.35 & 42.76 & 69.54 & 39.16 \\
& HQA &
84.43 & 51.23 & 65.83 & 50.25 & 80.99 & 64.89 & 48.91 & 38.24 & 65.77 & 36.53 & 83.86 & 50.50 \\
& NQ &
85.45 & 50.81 & 66.65 & 62.88 & 64.00 & 78.78 & 55.43 & 47.29 & 66.39 & 47.80 & 83.43 & 51.15 \\
\midrule
\multicolumn{3}{l}{\bf Multi-Task Learning}\\
\toprule
T5 E-Ext (B) & Multi &90.81 & 70.92 & 74.22 & 80.42 & 79.03 & 78.57 & 47.99 & 49.22 & 67.59 & 60.52 & 85.09 & 61.44 \\
T5 E-Ext (L) & Multi & 92.84 & 73.51 & 77.37 & 82.89 & 81.92 & 80.74 & 59.10 & 54.01 & 71.13 & 64.90 & 86.53 & 65.01 \\
T5 ED-Ext (B)& Multi & 91.12 & 71.95 & 75.50 & 81.82 & 80.25 & 78.93 & 49.69 & 49.64 & 68.45 & 61.33 & 85.49 & 61.22 \\
T5 ED-Gen (B)& Multi& 91.41 & 71.29 & 80.01 & 86.46 & 79.70 & 78.09 & 51.20 & 49.66 & 68.72 & 62.90 & 85.84 & 63.76 \\
BART E-Ext (L)& Multi & 92.46 & 72.11 & 72.24 & 76.53 & 82.04 & 79.40 & 58.22 & 50.40 & 70.72 & 56.29 & 86.79 & 61.95 \\
BART ED-Ext (L)& Multi & 93.07 & 58.67 & 71.47 & 80.66 & 82.14 & 79.32 & 60.40 & 51.32 & 67.48 & 53.34 & 86.75 & 56.79 \\
BART ED-Gen (L) & Multi&  88.63 & 68.91 & 74.91 & 82.52 & 80.53 & 75.78 & 55.20 & 50.04 & 63.78 & 54.81 & 80.94 & 58.47 \\
RoBERTa (L) & 93.41 & 59.56 & 72.23 & 80.98 & 82.37 & 79.55 & 64.47 & 51.81 & 69.15 & 53.68 & 86.31 & 56.06 \\
ELECTRA (L) & Multi & 93.27 & 60.59 & 72.96 & 82.03 & 83.10 & 79.16 & 62.56 & 50.29 & 71.50 & 54.60 & 87.14 & 56.88 \\
\bottomrule
\end{tabular}
}
}
\caption{Evaluation by F1 score. TQA: TriviaQA; SQA:SearchQA; HQA:HotpotQA; NQ: NaturalQuestions; TbQA:TextbookQA; RE:RelationExtraction. For inference length, T5 use Full length, BART and RoBERTa use 1024 and ELECTRA use 512. }
\label{tab:apd_table_f1_best}
\end{table*}

\end{document}